\PassOptionsToPackage{unicode}{hyperref}
\PassOptionsToPackage{hyphens}{url}
\PassOptionsToPackage{dvipsnames,svgnames,x11names}{xcolor}
\documentclass[
  12pt]{article}

\usepackage{amsmath,amssymb}
\usepackage{iftex}

\usepackage{lastpage}
\usepackage{mathtools} % amsmath with fixes and additions
\usepackage{booktabs} % commands to create good-looking tables
\usepackage{amsmath,amssymb,amsthm}
\usepackage{bbm}
\usepackage{enumitem}
\usepackage{hyperref}
\usepackage{multirow}
\usepackage{algorithm}
\usepackage{algpseudocode} % algorithmicx
\usepackage{tikz}
\usetikzlibrary{arrows.meta, positioning, calc}
\usepackage{pgfplots}
\pgfplotsset{compat=1.18} 
\usepackage{kotex}
\usepackage{placeins}
\usepackage{microtype}
\DeclareMathOperator{\TV}{\mathrm{TV}}

\newtheorem{theorem}{Theorem}
\newtheorem{lemma}{Lemma}
\newtheorem{assumption}{Assumption}

\newtheorem{definition}{Definition}
\newtheorem{remark}{Remark}
\newtheorem{proposition}{Proposition}
\newtheorem{example}{Example}
\usepackage{tcolorbox}
\tcbuselibrary{skins}
\newtcolorbox{informalbox}[1][]{
  colback=gray!10,
  colframe=gray!60,
  coltitle=black,
  fonttitle=\bfseries,
  title={#1},
  sharp corners,
  boxrule=0.5pt,
  left=5pt, right=5pt, top=5pt, bottom=5pt
}
\ifPDFTeX
  \usepackage[T1]{fontenc}
  \usepackage[utf8]{inputenc}
  \usepackage{textcomp} % provide euro and other symbols
\else % if luatex or xetex
  \usepackage{unicode-math}
  \defaultfontfeatures{Scale=MatchLowercase}
  \defaultfontfeatures[\rmfamily]{Ligatures=TeX,Scale=1}
\fi
\usepackage{lmodern}
\ifPDFTeX\else  
\fi
\IfFileExists{upquote.sty}{\usepackage{upquote}}{}
\IfFileExists{microtype.sty}{% use microtype if available
  \usepackage[]{microtype}
  \UseMicrotypeSet[protrusion]{basicmath} % disable protrusion for tt fonts
}{}
\makeatletter
\@ifundefined{KOMAClassName}{% if non-KOMA class
  \IfFileExists{parskip.sty}{%
    \usepackage{parskip}
  }{% else
    \setlength{\parindent}{0pt}
    \setlength{\parskip}{6pt plus 2pt minus 1pt}}
}{% if KOMA class
  \KOMAoptions{parskip=half}}
\makeatother
\usepackage{xcolor}
\makeatletter
\ifx\paragraph\undefined\else
  \let\oldparagraph\paragraph
  \renewcommand{\paragraph}{
    \@ifstar
      \xxxParagraphStar
      \xxxParagraphNoStar
  }
  \newcommand{\xxxParagraphStar}[1]{\oldparagraph*{#1}\mbox{}}
  \newcommand{\xxxParagraphNoStar}[1]{\oldparagraph{#1}\mbox{}}
\fi
\ifx\subparagraph\undefined\else
  \let\oldsubparagraph\subparagraph
  \renewcommand{\subparagraph}{
    \@ifstar
      \xxxSubParagraphStar
      \xxxSubParagraphNoStar
  }
  \newcommand{\xxxSubParagraphStar}[1]{\oldsubparagraph*{#1}\mbox{}}
  \newcommand{\xxxSubParagraphNoStar}[1]{\oldsubparagraph{#1}\mbox{}}
\fi
\makeatother

\usepackage{longtable,booktabs,array}
\usepackage{calc} % for calculating minipage widths
\usepackage{etoolbox}
\makeatletter
\patchcmd\longtable{\par}{\if@noskipsec\mbox{}\fi\par}{}{}
\makeatother
\IfFileExists{footnotehyper.sty}{\usepackage{footnotehyper}}{\usepackage{footnote}}
\makesavenoteenv{longtable}
\usepackage{graphicx}
\makeatletter
\def\maxwidth{\ifdim\Gin@nat@width>\linewidth\linewidth\else\Gin@nat@width\fi}
\def\maxheight{\ifdim\Gin@nat@height>\textheight\textheight\else\Gin@nat@height\fi}
\makeatother
\setkeys{Gin}{width=\maxwidth,height=\maxheight,keepaspectratio}
\makeatletter
\def\fps@figure{htbp}
\makeatother

\makeatletter
\@ifpackageloaded{caption}{}{\usepackage{caption}}
\AtBeginDocument{%
\ifdefined\contentsname
  \renewcommand*\contentsname{Table of contents}
\else
  \newcommand\contentsname{Table of contents}
\fi
\ifdefined\listfigurename
  \renewcommand*\listfigurename{List of Figures}
\else
  \newcommand\listfigurename{List of Figures}
\fi
\ifdefined\listtablename
  \renewcommand*\listtablename{List of Tables}
\else
  \newcommand\listtablename{List of Tables}
\fi
\ifdefined\figurename
  \renewcommand*\figurename{Figure}
\else
  \newcommand\figurename{Figure}
\fi
\ifdefined\tablename
  \renewcommand*\tablename{Table}
\else
  \newcommand\tablename{Table}
\fi
}
\@ifpackageloaded{float}{}{\usepackage{float}}
\floatstyle{ruled}
\@ifundefined{c@chapter}{\newfloat{codelisting}{h}{lop}}{\newfloat{codelisting}{h}{lop}[chapter]}
\floatname{codelisting}{Listing}

\makeatother
\makeatletter
\@ifpackageloaded{caption}{}{\usepackage{caption}}
\@ifpackageloaded{subcaption}{}{\usepackage{subcaption}}
\makeatother

\ifLuaTeX
  \usepackage{selnolig}  % disable illegal ligatures
\fi
\usepackage[]{natbib}
\usepackage{bookmark}

\IfFileExists{xurl.sty}{\usepackage{xurl}}{} % add URL line breaks if available
\hypersetup{
  pdftitle={When Should an AI Workflow Release? Always-Valid Inference for Black-Box Generate-Verify Systems},
  pdfauthor={Young Hyun Cho; Will Wei Sun},
  pdfkeywords={large language models, agentic AI workflows, selective release, verifier calibration, sequential hypothesis testing},
  colorlinks=true,
  linkcolor={blue},
  filecolor={Maroon},
  citecolor={Blue},
  urlcolor={Blue},
  pdfcreator={LaTeX via pandoc}}

\newcommand{\anon}{1}

\begin{document}

\def\spacingset#1{\renewcommand{\baselinestretch}%
{#1}\small\normalsize} \spacingset{1}

%%%%%%%%%%%%%%%%%%%%%%%%%%%%%%%%%%%%%%%%%%%%%%%%%%%%%%%%%%%%%%%%%%%%%%%%%%%%%%
\if1\anon
{
  \title{\bf When Should an AI Workflow Release? \\ Always-Valid Inference for Black-Box Generate-Verify Systems}
\author
{
Young Hyun Cho\thanks{Department of Statistics, Purdue University. Email: cho472@purdue.edu.}\qquad 
Will Wei Sun\thanks{Mitch Daniels School of Business, Purdue University. Email: sun244@purdue.edu. Corresponding author.}
}
  \maketitle
} \fi

\if0\anon
{
  \title{\bf When Should an AI Workflow Release? \\ Always-Valid Inference for Black-Box Generate-Verify Systems}
  \author{}
  \date{}
  \maketitle

} \fi

\bigskip

\begin{abstract}
%{\color{blue}
LLM-enabled AI workflows increasingly produce outputs through iterative generate-evaluate-revise loops. Each iteration can improve the candidate, but it also creates a release decision: when to stop and output the current result? This raises a statistical challenge because deployment-time evaluator scores are adaptively generated and repeatedly monitored, yet the likelihood models or exchangeability assumptions typically used for calibration are unavailable.
We propose an always-valid release wrapper for existing generator-evaluator pipelines. The wrapper builds a hard-negative reference pool of high-scoring failures, calibrates deployment-time evaluator scores against this pool, and accumulates the resulting evidence with an e-process. This separates two roles: the reference pool turns black-box scores into conservative evidence, while the e-process provides validity under optional stopping. In theory, we show that a conservative reference pool yields finite-sample control of the probability of releasing on infeasible tasks, that is, tasks for which the given workflow is not capable of producing a reliable solution. We also characterize conditions under which the same conservative rule still achieves nontrivial release on feasible tasks. 
In an MBPP+ coding-agent case study, the wrapper reduces premature incorrect release relative to baseline stopping rules while still releasing on tasks for which the workflow repeatedly accumulates moderate supporting evidence.
%}
\end{abstract}
%As LLM-based reasoning systems increasingly operate through adaptive generate-verify loops, deciding when to stop and release an answer becomes a statistical challenge. We study this problem in settings where a model generates candidate responses over time and receives scores from an external verifier, but standard likelihood-based and exchangeability-based calibration are unavailable. We propose an always-valid release wrapper that uses a conservative hard-negative reference pool to calibrate arbitrary verifier scores into $p$-values and then accumulates evidence through an e-process. The procedure can be layered on top of an existing pipeline without retraining the underlying model and improves release reliability under repeated querying and imperfect verification. In a case study on the MBPP+ code-generation benchmark, we show that the proposed wrapper yields more reliable release decisions than heuristic stopping rules.
%\end{abstract}

\noindent%
\vfill

\newpage
%\spacingset{1.8} % JASA. DON'T change the spacing!
\baselineskip=25pt %JASA allows at most 26 lines per page. Do not change

%{\color{blue}
\section{Introduction}
\label{sec:intro}

%{\color{blue}
Large language models (LLMs) are increasingly used as components of AI workflows rather than as one-shot predictors. In these workflows, a model is called repeatedly, candidate outputs are evaluated or revised over multiple rounds, and later calls are conditioned on what has already happened in the interaction. Industry reports describe growing interest in AI systems that move beyond chat-style assistance toward more workflow-integrated forms of problem solving \citep{mckinsey2025_agentic_ai_explained,mckinsey2026_agentic_ai_scale}. Guidance from AI labs likewise emphasizes traces, task outcomes, and failure modes as central objects to monitor in multi-step AI systems \citep{openai2026_eval_best_practices,anthropic2026_demystifying_evals_agents}. These developments broaden the scope of statistical decision-making for AI systems, bringing into view the interaction trajectory itself and the stopping decision that determines whether an output is released.

A recurring structure is an adaptive generate-evaluate-revise loop. At step $t$, the system produces a candidate output and observes an imperfect signal about its quality. This signal may come from tests, checkers, execution feedback, or learned judges; its role is not to certify correctness, but to provide evidence that the workflow may use when deciding whether to continue or release. Additional rounds can improve the candidate, but they also consume tokens, latency, and compute, and practical coding-agent systems already track these costs and activity patterns \citep{anthropic2026_claude_code_costs,anthropic2026_claude_code_monitoring}. Thus the deployment problem is a stopping problem: the system must decide when the evidence accumulated along an adaptive trajectory is strong enough to justify releasing the current output.
%}

\begin{figure}[h]
\centering
\resizebox{0.8\textwidth}{!}{
\begin{tikzpicture}[
    >=Stealth,
    node distance=2.1cm,
    every node/.style={text=black},
    box/.style={draw=black!65, rounded corners=1.5mm, thick, fill=white, align=center, minimum width=2.45cm, minimum height=0.78cm, font=\footnotesize, text=black},
    smallbox/.style={draw=black!45, rounded corners=1.2mm, thick, fill=black!3, align=center, minimum width=1.75cm, minimum height=0.66cm, font=\footnotesize, text=black},
    arrow/.style={->, thick, black!75}
]
\node[smallbox] (input) {Input};
\node[box, right=1.45cm of input] (gen) {Generator\\optimizer};
\node[box, right=2.10cm of gen] (eval) {Evaluator\\verifier};
\node[smallbox, right=1.45cm of eval] (out) {Accepted\\output};

\draw[arrow] (input) -- (gen);
\draw[arrow] (gen) -- node[above, font=\footnotesize, text=black] {candidate} (eval);
\draw[arrow] (eval) -- node[above, font=\footnotesize, text=black] {accept} (out);
\draw[arrow] (eval.south) .. controls +(0,-1.05) and +(0,-1.05) .. node[below, font=\footnotesize, text=black] {feedback for revision} (gen.south);
\end{tikzpicture}
}
\caption{Evaluator-optimizer workflow. A generator proposes a candidate, an evaluator returns either an acceptance signal or feedback, and rejected outputs are revised through an iterative loop. Adapted from the evaluator-optimizer workflow in Anthropic's \emph{Building Effective Agents} \protect\citep{anthropic2024_building_effective_agents}.}
\label{fig:evaluator_optimizer}
\end{figure}
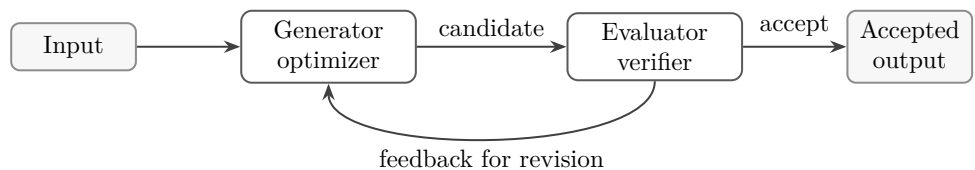

% %{\color{black}
% \begin{figure}[h]
% \centering
% \resizebox{0.8\textwidth}{!}{
% \begin{tikzpicture}[
%     >=Stealth,
%     node distance=2.1cm,
%     every node/.style={text=blue},
%     box/.style={draw=blue!65, rounded corners=1.5mm, thick, fill=white, align=center, minimum width=2.45cm, minimum height=0.78cm, font=\footnotesize, text=blue},
%     smallbox/.style={draw=blue!45, rounded corners=1.2mm, thick, fill=blue!3, align=center, minimum width=1.75cm, minimum height=0.66cm, font=\footnotesize, text=blue},
%     arrow/.style={->, thick, blue!75}
% ]
% \node[smallbox] (input) {Input};
% \node[box, right=1.45cm of input] (gen) {Generator\\optimizer};
% \node[box, right=2.10cm of gen] (eval) {Evaluator\\verifier};
% \node[smallbox, right=1.45cm of eval] (out) {Accepted\\output};

% \draw[arrow] (input) -- (gen);
% \draw[arrow] (gen) -- node[above, font=\footnotesize] {candidate} (eval);
% \draw[arrow] (eval) -- node[above, font=\footnotesize] {accept} (out);
% \draw[arrow] (eval.south) .. controls +(0,-1.05) and +(0,-1.05) .. node[below, font=\footnotesize] {feedback for revision} (gen.south);
% \end{tikzpicture}
% }
% \caption{\textbf{Evaluator-optimizer workflow.} A generator proposes a candidate, an evaluator returns either an acceptance signal or feedback, and rejected outputs are revised through an iterative loop. Adapted from the evaluator-optimizer workflow in Anthropic's \emph{Building Effective Agents} \protect\citep{anthropic2024_building_effective_agents}.}
% \label{fig:evaluator_optimizer}
% \end{figure}
% %}

%{\color{blue}
Figure~\ref{fig:evaluator_optimizer} abstracts a general generate-evaluate-revise workflow. This structure appears across mathematical reasoning with learned or process-level verifiers, coding workflows with execution feedback, and tool-using, web, robotic, or embodied agents that revise actions from external feedback \citep{cobbe2021training,lightman2023let,wang2024math,setlur2024rewarding,shao2024deepseekmath,guo2025deepseek,ma2025dynamic,hutter2026agentstepper,yao2022react,schick2023toolformer,ahn2022can,wang2023voyager,zhou2024webarena}. Across these domains, intermediate evaluator signals are useful but imperfect, so the release decision observes a black-box evidence stream rather than a certificate of correctness.

This adaptivity creates a statistical stopping problem. If a workflow repeatedly generates and evaluates candidates, then an incorrect or unreliable candidate receives repeated opportunities to obtain a favorable evaluator signal. A rule that simply waits for a high score can therefore be misled by repeated monitoring. This is the workflow analogue of $p$-hacking, where repeated looks at noisy evidence can inflate false discoveries unless the stopping rule is accounted for \citep{ramdas2025hypothesis}. The relevant statistical question is therefore not whether a candidate has a large raw score at a single step, but whether the evidence accumulated along the adaptive trajectory is strong enough to support release.

We treat this as a black-box release problem for adaptive workflows. The wrapper does not retrain the generator or evaluator, nor model the distribution of workflow traces. It adds a distribution-free decision layer using verifier signals observed during deployment. The goal is to control false release on instances where the given pipeline cannot produce a reliable solution, while still allowing release when repeated evidence supports feasibility.
%}

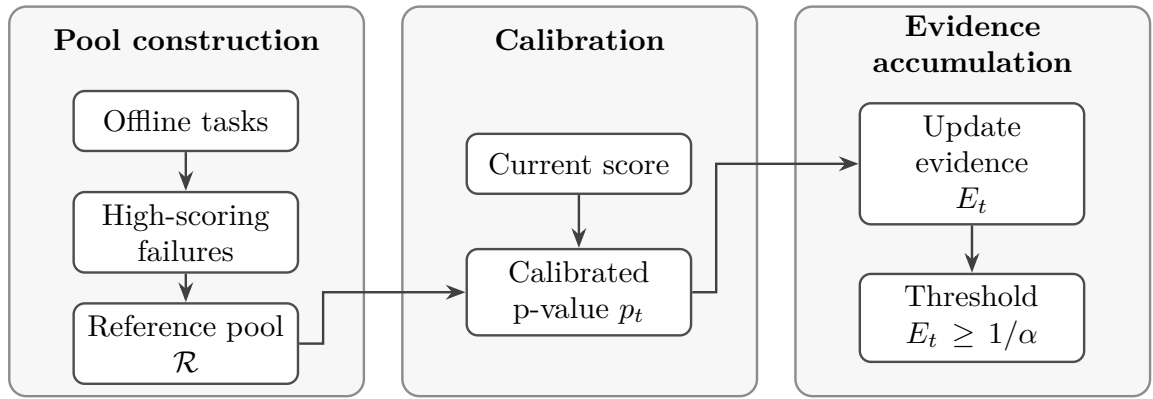
\begin{figure}[t]
\centering
\resizebox{0.95\textwidth}{!}{
\begin{tikzpicture}[
    >=Stealth,
    every node/.style={text=black},
    stage/.style={
        draw=black!45,
        rounded corners=2mm,
        thick,
        fill=black!3,
        minimum height=4.55cm
    },
    title/.style={
        font=\bfseries\footnotesize,
        align=center,
        text=black
    },
    box/.style={
        draw=black!70,
        rounded corners=1.3mm,
        thick,
        fill=white,
        align=center,
        minimum height=7mm,
        text width=2.35cm,
        font=\footnotesize,
        text=black
    },
    arrow/.style={->, thick, black!75}
]

% Stage containers
\node[stage, minimum width=4.15cm, anchor=north west] (S1) at (0,0) {};
\node[stage, minimum width=4.15cm, anchor=north west] (S2) at (4.60,0) {};
\node[stage, minimum width=4.15cm, anchor=north west] (S3) at (9.20,0) {};

\node[title] at ($(S1.north)+(0,-0.40)$) {Pool construction};
\node[title] at ($(S2.north)+(0,-0.40)$) {Calibration};
\node[title] at ($(S3.north)+(0,-0.44)$) {Evidence\\accumulation};

% Stage 1
\node[box] (offline) at ($(S1.north west)+(2.08,-1.35)$) {Offline tasks};
\node[box] (fails)   at ($(S1.north west)+(2.08,-2.65)$) {High-scoring\\failures};
\node[box] (pool)    at ($(S1.north west)+(2.08,-3.95)$) {Reference pool\\\(\mathcal R\)};
\draw[arrow] (offline) -- (fails);
\draw[arrow] (fails) -- (pool);

% Stage 2
\node[box] (score) at ($(S2.north west)+(2.08,-1.85)$) {Current score};
\node[box] (pt)    at ($(S2.north west)+(2.08,-3.35)$) {Calibrated\\p-value \(p_t\)};
\draw[arrow] (score) -- (pt);
\draw[arrow] (pool.east) -- ++(0.28,0) |- (pt.west);

% Stage 3
\node[box] (Et)   at ($(S3.north west)+(2.08,-1.85)$) {Update evidence\\\(E_t\)};
\node[box] (rule) at ($(S3.north west)+(2.08,-3.65)$) {Threshold\\\(E_t \ge 1/\alpha\)};
\draw[arrow] (pt.east) -- ++(0.28,0) |- (Et.west);
\draw[arrow] (Et) -- (rule);

\end{tikzpicture}
}
\caption{Overview of the proposed release wrapper. An offline collection of tasks is used to construct a reference pool \(\mathcal R\). During online deployment, the current evaluator score is calibrated against \(\mathcal R\) to obtain a stepwise \(p\)-value \(p_t\), which is then accumulated into sequential evidence \(E_t\). Release occurs only when \(E_t\) crosses the threshold \(1/\alpha\).}
\label{fig:three_stage_wrapper}
\end{figure}

Figure~\ref{fig:three_stage_wrapper} shows the decision layer proposed in this paper. The hard-negative reference pool supplies a failure-side calibration anchor: a deployment-time score is not judged by its raw magnitude, but by how extreme it is relative to scores previously achieved by high-scoring failures. This converts heterogeneous black-box evaluator signals into calibrated stepwise \(p\)-values. An e-process then accumulates these calibrated values over time and releases only after the accumulated evidence crosses an anytime-valid threshold. 
%}

\subsection{Our Contributions}

%{\color{blue}
Our contributions can be summarized as follows:

\noindent\textbf{Statistical formulation of safe release in adaptive workflow.} We formulate verifier-guided adaptive workflows as a selective release problem, with safety measured by \(\mathbb{P}(D=1\mid F=0)\), the probability of release on instances for which the specified generator-verifier pipeline is not capable of producing a reliable solution. This casts deployment-time stopping as a statistical control problem rather than a heuristic rule for black-box verifier scores.

\noindent\textbf{Methodology.}
We introduce a modular wrapper for black-box generator-evaluator workflows. The wrapper does not retrain either component or model the distribution of workflow traces. It builds a conservative hard-negative reference pool from offline failures, calibrates deployment-time verifier scores by upper-tail ranks against this pool, and accumulates the calibrated evidence through an e-process. This modular design is useful for rapidly changing AI systems because updates to the generator, evaluator, or task distribution can be handled by rebuilding the reference pool, while the release rule remains unchanged.

\noindent\textbf{Theory.}
The main theoretical issue is how to construct valid sequential testing from adaptive black-box verifier scores. Likelihood-ratio e-processes require conditional score laws, while conformal rank arguments require exchangeability; neither structure is naturally available in our problem. We introduce a failure-side calibration dominance condition under which empirical upper-tail ranks become conditionally super-uniform \(p\)-values despite history-dependent score generation. Combined with $p$-to-e accumulation, this yields \(\mathbb{P}(D=1\mid F=0)\le\alpha\). We also give a finite-horizon result showing that our wrapper can release on feasible tasks when correct candidates provide enough calibrated evidence over time.

\noindent\textbf{Empirics.}
We use MBPP+ code generation \citep{liu2023your} as a coding-agent instance of the broader release problem. The case study shows that baseline stopping rules based on confidence or score stability can release too early, while one-step calibrated thresholding can become too conservative when no single verifier score is decisive. In contrast, the proposed wrapper accumulates moderate evidence across iterations, reducing premature incorrect release while still releasing on many tasks that the underlying workflow eventually solves.
%}

\subsection{Related Work}
%{\color{blue}
Recent work across LLM-enabled workflows increasingly relies on iterative generation with external or learned feedback. In mathematical reasoning, trained verifiers, process reward models, and test-time search guide candidate solutions \citep{cobbe2021training,lightman2023let,setlur2024rewarding,wang2024math,shao2024deepseekmath,guo2025deepseek}. In software and agentic settings, execution results, tool outputs, browser states, or environment feedback play a similar role in guiding revision \citep{ma2025dynamic,hutter2026agentstepper,yao2022react,schick2023toolformer,madaan2023self,shinn2023reflexion,ahn2022can,wang2023voyager,zhou2024webarena}. When deployment-time stopping is addressed, it is often handled by heuristics based on confidence, entropy, stability, or related proxies for sufficient progress \citep{li2024escape,wu2025efficiency,mao2025early,laaouach2025halt,quamar2025logit,wu2025thought,huang2026does}. By contrast, we treat release as a statistical control problem for a black-box workflow and develop a wrapper that can be layered on top of an existing pipeline without retraining either component.
%}
Our setting is also related to \citet{xie2026statistical}, which develops stopping rules from signals from internal reasoning traces. The distinction is that their approach relies on white-box access to internal markers of the reasoning process, whereas we assume no access to internal states and operate only through external verifier scores. 

Methodologically, our work connects to always-valid inference and e-processes \citep{ramdas2025hypothesis, su2026online}, which provide time-uniform validity once a valid stream of stepwise evidence is available. In our problem, however, the main difficulty arises earlier: the deployment-time verifier scores are adaptive, heterogeneous, and generated by a black-box workflow, so they are not themselves valid e-values or \(p\)-values. Our contribution is therefore not the use of e-processes alone, but the construction of a valid testing interface for these scores. Moreover, our rank-calibration step is related to conformal prediction \citep{vovk2005algorithmic, barber2023conformal}, because both approaches compare a new score with a reference collection. The source of validity, however, is different. Standard conformal guarantees rely on exchangeability between the calibration sample and the new observation, whereas the adaptive score process generated by a black-box evaluator--optimizer loop generally admits neither exchangeability nor a tractable likelihood model. We address this by constructing a pool of hard-negative failures that serves as a conservative failure-side anchor. Under a one-sided dominance condition, this reference pool yields stepwise-valid \(p\)-values, which can then be accumulated through the e-process to obtain always-valid release decisions.

% {\color{blue}
% Methodologically, our work connects to always-valid inference and e-processes \citep{ramdas2025hypothesis, su2026online}, which provide time-uniform validity once valid evidence is available. In our problem, however, the central difficulty arises one step earlier: the online verifier scores are not already valid e-values. On the other hand, our rank-calibration step is related in spirit to conformal prediction \citep{vovk2005algorithmic, barber2023conformal}, because both approaches compare a score to a reference collection. However, the source of validity is different. Standard conformal guarantees rely on exchangeability, whereas the adaptive score process generated by a black-box evaluator-optimizer loop generally admits neither exchangeability nor a tractable likelihood model. We address this by constructing a pool of hard-negative failures that serves as a conservative failure-side anchor for obtaining stepwise-valid \(p\)-values, which are then accumulated through the e-process.
% }
\section{Problem Setup and Preliminaries}
\label{sec:setup}

{\color{black}
We study a sequential release problem for an input instance \(x\in\mathcal X\), drawn from a task distribution \(P_X\). The input may be a programming task, a mathematical problem, or another task handled by an adaptive AI workflow. A generator produces a sequence of candidate outputs, and after each candidate an evaluator or verifier returns a scalar score summarizing current evidence for correctness. The wrapper observes this evolving interaction and must decide whether to release the current candidate or continue searching. %In a coding workflow, the generator may revise a program based on execution feedback, while the score may be the fraction of visible unit tests passed, or an LLM-based code judge. In a mathematical or tool-using workflow, the score may come from answer matching, symbolic checking, tool feedback, or an LLM judge evaluating plausibility. In all cases, the interaction is adaptive: the next candidate depends on the history, and the evaluator itself may be a black-box. We therefore impose no specific parametric assumptions on the generator-evaluator interaction.
}

Formally, at each step \(t\ge1\), the generator proposes a candidate output \(C_t\), and the verifier returns a scalar signal \(S_t\in\mathcal S\subseteq\mathbb R\). Let
\[
H_t := (x,C_1,S_1,\ldots,C_t,S_t)
\]
denote the observable interaction history up to step \(t\), and let
$
\mathcal G_t := \sigma(H_t)
$
be the corresponding filtration. The wrapper uses \(H_t\) to decide whether to release the current candidate. Let \(D_t\in\{0,1\}\) be the step-\(t\) decision, where \(D_t=1\) means \emph{release now} and \(D_t=0\) means \emph{continue}. We require \(D_t\) to be \(\mathcal G_t\)-measurable. The release time is
\[
\tau := \inf\{t\ge1:D_t=1\},
\]
with the convention \(\tau=\infty\) if the wrapper never releases. Given a maximum horizon \(T_{\max}\), the system releases \(C_\tau\) if \(\tau\le T_{\max}\) and abstains otherwise. We write
\[
D := \mathbf 1\{\tau\le T_{\max}\}
\]
for the terminal release indicator.

Let \(Y_t\in\{0,1\}\) denote whether candidate \(C_t\) is truly correct. Thus \(Y_t\) is a task-specific correctness label: in coding, \(Y_t=1\) may mean that the generated program passes the full hidden test suite; in math reasoning, \(Y_t=1\) may mean that the final answer is mathematically correct. The wrapper does not observe \(Y_t\) online; it observes only the verifier score \(S_t\), which may be informative but imperfect.

We also introduce a latent feasibility state
\[
F\in\{0,1\}.
\]
The variable \(F\) distinguishes two qualitatively different reasons that a candidate may be incorrect. A candidate may be wrong simply because the search is still ongoing, even though the underlying generator-verifier pipeline is capable of eventually producing a reliable solution. Alternatively, the instance may be effectively beyond the capability of the given pipeline, so that even high verifier scores do not correspond to genuine eventual success. We interpret \(F=1\) as the \emph{feasible} regime and \(F=0\) as the \emph{infeasible} regime. Importantly, \(F\) is a property of the entire pipeline applied to the instance, not of the instance alone.

This distinction is useful in adaptive workflows. In coding, an incorrect program may still pass many visible tests, so a high \(S_t\) can reflect overfitting rather than true correctness. In mathematical reasoning, an incorrect derivation may appear polished and receive a favorable judge score even though the problem is beyond the model's effective capability. The wrapper should not be penalized for early incorrect candidates on feasible tasks, but it should control the probability of release on infeasible tasks.

\begin{example}[Coding and math reasoning]
\leavevmode
\begin{itemize}[leftmargin=1.5em, itemsep=0.15em, topsep=0.15em]
    \item \textbf{Coding.} \(x\) is a programming problem, \(C_t\) is a proposed program, and \(S_t\) may be the fraction of visible unit tests passed. The label \(Y_t=1\) means that the program is actually correct, for example because it passes the full hidden test suite. The state \(F=1\) means that the specified generator-verifier pipeline can in principle produce a reliable solution for this task; \(F=0\) means that it cannot, even though some intermediate programs may achieve deceptively high verifier scores.
    
    \item \textbf{Math reasoning.} \(x\) is a mathematical problem, \(C_t\) is a reasoning trace together with a final answer, and \(S_t\) may come from exact answer matching, symbolic checking, or an LLM judge. The label \(Y_t=1\) means that the final answer is mathematically correct. The state \(F=1\) means that the pipeline can in principle reach a reliable solution for this problem; \(F=0\) means that it cannot, even if some candidate derivations appear plausible or receive high verifier scores.
\end{itemize}
\end{example}

The release decision can be viewed through the lens of hypothesis testing. The infeasible regime \(F=0\) plays the role of a null state, while the feasible regime \(F=1\) plays the role of an alternative state. Releasing a candidate, that is, deciding \(D=1\), is then analogous to rejecting the null. %This analogy is only conceptual, since our setting is sequential, adaptive, and likelihood-free, but it provides a useful way to organize the relevant operating characteristics.
This perspective leads to two primitive criteria. The first is the false-release probability on infeasible instances,
\begin{equation}
\mathbb P(D=1\mid F=0),
\label{eq:false_release_infeasible}
\end{equation}
which measures how often the wrapper is misled into release when the pipeline is not capable of producing a reliable solution. This quantity plays the role of a Type~I error analogue.

The second is the release probability on feasible instances,
\begin{equation}
\mathbb P(D=1\mid F=1),
\label{eq:release_feasible}
\end{equation}
which measures how often the wrapper releases when a reliable solution is attainable. This quantity plays the role of power. A wrapper that always abstains would trivially drive \eqref{eq:false_release_infeasible} to zero, but only at the cost of also setting \(\mathbb P(D=1\mid F=1)=0\).

Accordingly, our main safety target is
\begin{equation}
\mathbb P(D=1\mid F=0)\le \alpha,
\label{eq:primary_guarantee}
\end{equation}
which is the analogue of Type~I error control at level \(\alpha\).

The most user-facing notion of reliability is the release-conditional failure rate
\begin{equation}
R_{\mathrm{out}}
:=
\mathbb P(Y_\tau=0\mid D=1),
\label{eq:rout_def}
\end{equation}
which is the probability that a released candidate is incorrect. This is the ultimate practical target, but it is difficult to control directly because the released candidate is selected through a history-dependent stopping rule and the wrapper observes only verifier scores rather than ground-truth correctness. For this reason, our theory focuses first on the more tractable target \eqref{eq:primary_guarantee}, together with the companion feasible-side release probability \eqref{eq:release_feasible}.

The next proposition links the primitive safety target to the release-conditional failure rate.

\begin{proposition}
\label{prop:rout_from_primitive}
Let \(\pi_0=\mathbb P(F=0)\). Suppose the wrapper satisfies
\(\mathbb P(D=1\mid F=0)\le \alpha\).
Then
\begin{equation}
\begin{split}
R_{\mathrm{out}}
&\le
\mathbb P(Y_\tau=0\mid D=1,F=1) \\
&\quad+
\Bigl(1-\mathbb P(Y_\tau=0\mid D=1,F=1)\Bigr)
\frac{\pi_0\alpha}{\pi_0\alpha+(1-\pi_0)\mathbb P(D=1\mid F=1)}.
\end{split}
\label{eq:rout_translation}
\end{equation}
\end{proposition}

The decomposition in \eqref{eq:rout_translation} highlights the division of roles between the underlying reasoning pipeline and the release wrapper. The first term reflects performance on feasible instances and is determined by the quality of the base generator once the wrapper releases. The second term isolates the contribution from infeasible instances and is controlled directly through the false-release bound \(\alpha\).

The primitive safety target \eqref{eq:primary_guarantee} requires tools from sequential inference. A natural baseline is to compute, at each step \(t\), a significance measure based on the current verifier signal, for example, a \(p\)-value that would be valid if the analysis stopped at time \(t\) alone, and to release as soon as this quantity falls below a nominal threshold. While intuitive, this strategy is fundamentally unreliable under repeated monitoring.

\begin{proposition}[Failure of naive stopping]
\label{thm:optstop_ct}
Fix \(\alpha\in(0,1)\) and let \((p_t)_{t\ge1}\) be a sequence of \([0,1]\)-valued statistics adapted to a filtration \((\mathcal G_t)_{t\ge1}\). Define
$
\tau_{\mathrm{naive}}:=\inf\{t\ge1:p_t\le \alpha\}.
$
If there exists a deterministic sequence \((c_t)_{t\ge1}\subset[0,1]\) such that for all \(t\ge1\),
$
\mathbb P(p_t\le\alpha\mid \mathcal G_{t-1},F=0)\ge c_t
$
almost surely, then for any \(T\in\mathbb N\),
\[
\mathbb P(\tau_{\mathrm{naive}}\le T\mid F=0)
\ge
1-\exp\!\left(-\sum_{t=1}^T c_t\right).
\]
Consequently, if \(\sum_{t\ge1} c_t=\infty\), then
$
\mathbb P(\tau_{\mathrm{naive}}<\infty\mid F=0)=1.
$
\end{proposition}

Proposition~\ref{thm:optstop_ct} shows that repeated thresholding of stepwise significance measures does not yield sequential safety. Even if each individual step has only a tiny chance of producing a misleadingly favorable statistic, repeated monitoring causes these risks to accumulate over time. In our setting, this is the sequential analogue of ``\(p\)-hacking''.

To avoid this failure, we need a notion of evidence that remains valid under arbitrary stopping times. E-processes provide exactly this type of time-uniform control.

\begin{definition}[e-process]
Fix a null hypothesis \(H_0\). Let \((\mathcal H_t)_{t\ge0}\) be a filtration. A nonnegative \((\mathcal H_t)\)-adapted process \((E_t)_{t\ge0}\) is an \emph{e-process} for \(H_0\) if \(E_0=1\) and, under \(H_0\),
\begin{equation}
\label{equ:eprocess_def}
\mathbb E_{H_0}[E_t\mid \mathcal H_{t-1}] \le E_{t-1}
\qquad
\text{almost surely for all } t\ge1.
\end{equation}
\end{definition}
In our application, the null hypothesis corresponds to the infeasible regime \(F=0\). The following classical inequality makes the time-uniform validity of an e-process precise.

\begin{lemma}[Ville's inequality \citep{doob1939jean}]
\label{lem:ville}
If \((E_t)_{t\ge0}\) is an e-process for \(H_0\), then
\[
\mathbb P_{H_0}\!\left(\sup_{t\ge1} E_t\ge \frac{1}{\alpha}\right)\le \alpha, 
\]
for any \(\alpha\in(0,1)\). Equivalently, for
$
\tau_\alpha:=\inf\left\{t\ge1:E_t\ge \frac{1}{\alpha}\right\},
$
we have
$
\mathbb P_{H_0}(\tau_\alpha<\infty)\le \alpha.
$
\end{lemma}

Lemma~\ref{lem:ville} implies that constructing an e-process is sufficient for always-valid safe release. In classical sequential testing, e-processes are often derived from likelihood ratios \citep{ramdas2023game}. Our setting, however, is black-box and likelihood-free: the verifier scores are heterogeneous, adaptively generated, and need not admit a tractable parametric model. A robust alternative is \emph{p-to-e calibration} \citep{vovk2021values}, which transforms a sequence of valid \(p\)-values into an e-process. The main challenge, therefore, is to construct valid \(p\)-values from the raw verifier scores \(S_t\). %Our solution is to use a conservative hard-negative reference pool as a failure benchmark. The next section develops this construction and shows how it yields always-valid release decisions in adaptive reasoning pipelines.

{\color{black}
\section{Proposed Method and Its Statistical Properties}
\label{sec:method}

In this section, we construct the failure-calibrated release wrapper and explain how it provides a statistical testing interface for black-box verifier scores.% We first describe the offline hard-negative reference pool that anchors the failure-side calibration. We then formalize the dominance condition under which raw, history-dependent verifier scores can be converted into conditionally valid empirical upper-tail \(p\)-values. Finally, we combine these calibrated \(p\)-values with an e-process to obtain an always-valid release rule, and we analyze feasible-side power and finite-horizon calibration drift. The emphasis is that the e-process is the optional-stopping engine, while the hard-negative calibration layer is what turns adaptive LLM verifier outputs into valid inputs for sequential testing.

\subsection{Algorithm overview}
\label{subsec:method_overview}

At each interaction step \(t\), the system produces a candidate output \(C_t\) and a verifier score \(S_t\). The wrapper proceeds in three stages.
\begin{enumerate}[leftmargin=1.5em, itemsep=-0.5em, topsep=0.2em]
    \item \textbf{Offline hard-negative reference-pool construction.} Using an auxiliary calibration collection, construct a pool \(\mathcal R\) of incorrect candidates that nevertheless receive unusually favorable verifier scores.
    \item \textbf{Reference-pool calibration.} Compare the current score \(S_t\) to the fixed pool \(\mathcal R\) and convert it into an empirical upper-tail \(p\)-value \(p_t\).
    \item \textbf{Sequential evidence accumulation and release.} Transform the calibrated \(p\)-values into e-values, accumulate them over time, and release only when the resulting evidence process crosses \(1/\alpha\).
\end{enumerate}
The wrapper acts on rank-calibrated evidence rather than on raw verifier scores. A score that appears large in isolation may not be compelling if similar scores are frequently produced by incorrect outputs. A score that remains extreme even relative to a pool of hard negatives is more meaningful evidence against the infeasible regime.

\subsection{Offline hard-negative reference-pool construction}
\label{subsec:reference_pool}

The first step is to construct a conservative reference pool \(\mathcal R=\{R_1,\ldots,R_n\}\), which will later be used to calibrate online verifier scores. The need for such a pool comes from the fact that the raw score $S_{t}$ is often heterogeneous across tasks and difficult to interpret on an absolute scale. In code generation, for example, a score of 0.8 may mean that the candidate passes 80\% of a visible test suite, but whether that is impressive depends on the task and on how easy the visible tests are to overfit. In mathematical reasoning, a high judge score may simply reflect that a derivation looks plausible even if it is incorrect. We therefore build the reference pool from hard negatives, namely incorrect candidates that nevertheless receive favorable verifier scores, and evaluate each online score through its relative extremeness against this benchmark.

\begin{algorithm}[t]
\caption{Offline construction of a hard-negative reference pool}
\label{alg:reference_pool}
\begin{algorithmic}[1]
\Require Auxiliary calibration set \(\mathcal D_{\mathrm{cal}}\), generator \(\mathsf{Gen}\), online verifier \(\mathsf{Ver}\), stronger offline adjudicator \(\mathcal A\), upper-tail selection rule \(\Gamma\)
\Ensure Fixed hard-negative reference pool \(\mathcal R=\{R_1,\dots,R_n\}\)

\State Initialize incorrect-score list \(\mathcal S_{\mathrm{inc}} \gets [\,]\)
\For{each calibration instance \(x \in \mathcal D_{\mathrm{cal}}\)}
    \State Generate one or more candidate outputs \(C\) using \(\mathsf{Gen}(x)\)
    \For{each generated candidate \(C\)}
        \State Compute verifier score \(s \gets \mathsf{Ver}(x,C)\)
        \If{\(\mathcal A(x,C)=0\)}
            \State Append \(s\) to \(\mathcal S_{\mathrm{inc}}\)
        \EndIf
    \EndFor
\EndFor
\State Apply the upper-tail selection rule \(\Gamma\) to \(\mathcal S_{\mathrm{inc}}\)
\State Retain the selected high-scoring incorrect values to form \(\mathcal R\)
\State Optionally validate candidate pools on held-out calibration data and refine \(\Gamma\) if needed
\State \Return \(\mathcal R\)
\end{algorithmic}
\end{algorithm}

Formally, let \(\mathcal D_{\mathrm{cal}}\) denote a calibration set drawn from the same task family. For each \(x\in\mathcal D_{\mathrm{cal}}\), we generate candidate output \(C\), compute its online verifier score \(s=\mathsf{Ver}(x,C)\), and then apply a stronger offline adjudicator \(\mathcal A(x,C)\in\{0,1\}\), where \(\mathcal A(x,C)=1\) means that the candidate is judged correct by the stronger offline criterion. We retain only those candidates with \(\mathcal A(x,C)=0\).

Among the incorrect candidates, we retain only those whose scores lie in the upper tail of the failure distribution. This restriction is important because a pool built from arbitrary failures need not be conservative for selective release. If many reference scores come from obviously poor candidates with very low verifier values, then even a moderately favorable online score may appear artificially extreme, producing anti-conservative calibrated \(p\)-values. Hard-negative selection instead enlarges the right tail of the failure-side reference distribution, focusing calibration on the score region most relevant to false release.

To make this selection explicit, let \(\{\Gamma_q:q\in(0,1]\}\) be a family of upper-tail selection rules, where \(\Gamma_q\) retains the upper \(100q\%\) of incorrect-candidate scores. In the MBPP+ case study, \(\Gamma_q\) is applied to visible verifier scores among incorrect bank-split candidates. In particular, \(\Gamma_{0.55}\) yields the \texttt{top55} pool used in the main analysis of the experiments.

After constructing a candidate pool, we compute the induced empirical upper-tail \(p\)-values on held-out incorrect examples and examine their empirical CDF
\[
\widehat F(u):=\widehat{\mathbb P}(p_t\le u), \qquad u\in[0,1].
\]
A conservative pool should yield an empirical CDF lying at or below the diagonal \(u\mapsto u\). This diagnostic provides a practical screen for ruling out visibly anti-conservative candidate pools before deployment.

%Although this diagnostic does not verify the full history-conditional dominance condition required for exact sequential validity, it provides a practical screen for ruling out visibly anti-conservative candidate pools before deployment.

%The reference pool is therefore not only an implementation device. Its statistical role appears in the next subsection. It provides the failure-side anchor that replaces the likelihood specification or exchangeability assumptions used by more standard sequential testing and conformal constructions.

\subsection{Reference-pool calibration and e-process construction}
\label{subsec:calibration}

With the offline hard-negative pool \(\mathcal R\) fixed, we now construct the testing used during deployment. Let \((\mathcal G_t)_{t\ge0}\) denote the observable filtration generated by the online interaction, and define \(\widetilde{\mathcal G}_t:=\mathcal G_t\vee\sigma(\mathcal R)\), which additionally reveals the realized reference pool.

The main obstacle is not optional stopping itself. Once a valid evidence stream is available, e-processes provide a standard mechanism for accumulating it under arbitrary stopping. The obstacle is constructing that stream from raw verifier scores. The online verifier score \(S_t\) is produced by an adaptive black-box interaction, so its conditional law is history-dependent and typically unavailable. As a result, standard likelihood-based or exchangeability-based calibration cannot be applied directly. Specifically, a likelihood-ratio route would require conditional score laws under the feasible and infeasible regimes. If such laws were available, say \(q_{1,t}(\cdot\mid\mathcal G_{t-1})\) and \(q_{0,t}(\cdot\mid\mathcal G_{t-1})\), one could form the likelihood-ratio evidence process \(E_t^{\mathrm{LR}}:=\prod_{j=1}^t q_{1,j}(S_j\mid\mathcal G_{j-1})/q_{0,j}(S_j\mid\mathcal G_{j-1})\). In our problem, these conditional laws are not specified. The generator-verifier interaction is black-box, and the candidate \(C_t\) is adapted to the previous history. Thus neither \(q_{0,t}\) nor \(q_{1,t}\) is available as an input to the release rule.

A conformal perspective is useful because our calibration step also compares an online score with a reference collection. If \(R_1,\ldots,R_n\) and \(S_t\) were exchangeable under the infeasible regime, the usual rank argument would yield a valid \(p\)-value by symmetry. The key difference is the source of validity. In an adaptive generator-verifier loop, the candidate \(C_t\) is produced after observing the previous interaction, so the conditional law of \(S_t\) may change with \(\mathcal G_{t-1}\). Thus the online score stream is not exchangeable with a fixed offline calibration sample. The hard-negative pool is therefore not used as an exchangeable conformal calibration set. It is used as a conservative failure-side anchor, and validity will come from the following one-sided dominance condition defined in \eqref{eq:dom_tail} rather than from exchangeability.

%The hard-negative pool provides a third route. Instead of specifying a likelihood or assuming exchangeability, we require a one-sided failure-side dominance condition. 
\begin{assumption}[Failure-side calibration dominance]
\label{assump:dom}
Under the infeasible regime \(F=0\), the reference pool stochastically dominates the online score generation. Specifically, for every step \(t \ge 1\) and threshold \(s \in \mathbb R\),
\begin{equation}
\mathbb P\!\left(S_t \ge s \,\middle|\, \widetilde{\mathcal G}_{t-1}, F=0\right)
\le
\mathbb P\!\left(\tilde S \ge s \,\middle|\, \sigma(\mathcal R)\right),
\label{eq:dom_tail}
\end{equation}
almost surely, where \(\tilde S\) is an independent draw from the empirical distribution supported on \(\mathcal R\), conditional on \(\sigma(\mathcal R)\).
\end{assumption}

Assumption~\ref{assump:dom} is the calibration interface between black-box LLM scores and sequential testing. It does not model the full conditional law of \(S_t\), nor does it assume that online and offline scores are exchangeable. It only requires that, under \(F=0\), the online upper tail is no heavier than the upper tail represented by the hard-negative pool. Algorithm~\ref{alg:reference_pool} is designed to make this condition plausible by using the same generator-verifier pipeline as deployment, filtering with a stronger offline adjudicator, and retaining high-scoring failures.

Given \(\mathcal R\), define the empirical upper-tail \(p\)-value
\begin{equation}
p_t := \frac{1 + \sum_{i=1}^n \mathbf{1}\{R_i \ge S_t\}}{n+1}.
\label{eq:tail_pvalue}
\end{equation}
Thus \(p_t\) is small when the current online score \(S_t\) is large relative to the hard-negative benchmark. The floor \(1/(n+1)\) prevents zero \(p\)-values when \(S_t\) exceeds all reference scores.

\begin{theorem}[Stepwise validity under a conservative reference pool]
\label{thm:stepwise_pvalue}
Under Assumption~\ref{assump:dom}, for every \(u \in [0,1]\) and every \(t \ge 1\), it holds almost surely that
\begin{equation}
\mathbb P\!\left(p_t \le u \,\middle|\, \widetilde{\mathcal G}_{t-1}, F=0\right)
\le u.
\label{eq:su}
\end{equation}
\end{theorem}

Theorem~\ref{thm:stepwise_pvalue} is the point at which a raw adaptive verifier score becomes a testing object. Before Theorem~\ref{thm:stepwise_pvalue}, \(S_t\) is only a black-box score whose scale and distribution are task-dependent and history-dependent. After Theorem~\ref{thm:stepwise_pvalue}, the rank-calibrated quantity \(p_t\) is conditionally super-uniform under infeasibility and can therefore be used as valid sequential testing input. This is the main technical role of the hard-negative reference pool.

%The result also clarifies the division of labor in the wrapper. 
The result also clarifies the respective roles of the two components of the wrapper.
The hard-negative calibration step supplies validity for the transformation \(S_t\mapsto p_t\), while the e-process supplies optional-stopping validity for the transformation \(p_t\mapsto E_t\). The contribution is the construction of an interface for adaptive black-box verifier scores, which then allows standard anytime-valid accumulation to be used.

We next convert the calibrated \(p\)-values into an e-process. This step is standard once Theorem~\ref{thm:stepwise_pvalue} has supplied the valid input stream. Let \(f_t:[0,1]\to[0,\infty)\) be a predictable non-increasing betting function satisfying \(\int_0^1 f_t(u)\,du \le 1\) almost surely for each \(t\). Define
\[
E_t := \prod_{s=1}^t f_s(p_s), \qquad E_0:=1.
\]

\begin{proposition}[E-process from calibrated \(p\)-values]
\label{prop:betting_martingale}
Suppose \((p_t)_{t\ge1}\) satisfies \eqref{eq:su}. Let \((f_t)_{t\ge1}\) be a sequence of predictable non-increasing functions with \(\int_0^1 f_t(u)\,du \le 1\) almost surely. Then, under \(F=0\), it holds almost surely for all \(t\ge1\) that
\[
\mathbb E\!\left[E_t \mid \widetilde{\mathcal G}_{t-1},F=0\right]
\le E_{t-1}.
\]
Hence \((E_t)_{t\ge0}\) is an e-process for the infeasible regime.
\end{proposition}

Proposition~\ref{prop:betting_martingale} is the optional-stopping component of the wrapper. Its role is to accumulate the conditionally valid \(p\)-values supplied by Theorem~\ref{thm:stepwise_pvalue}. %In this sense, the proof architecture separates the new calibration interface \(S_t\mapsto p_t\) from the standard p-to-e accumulation \(p_t\mapsto E_t\).

In the implementation, we use the standard truncated power-betting family \citep{vovk2021values, grunwald2020safe}. Specifically, for \(\eta\in(0,1)\) and truncation level \(M\ge1\), let \(f^{(\eta,M)}(u):=c\min(u^{-\eta},M)\), where \(c\) is chosen so that \(\int_0^1 f^{(\eta,M)}(u)\,du=1\).

We define the release time and terminal release indicator by
\[
\tau_\alpha := \inf\{t\ge1:E_t\ge 1/\alpha\},
\qquad
D:=\mathbf 1\{\tau_\alpha\le T_{\max}\}.
\]
That is, the wrapper releases only when the accumulated evidence against the infeasible regime exceeds \(1/\alpha\) within the maximum horizon \(T_{\max}\). Otherwise, it abstains. The detailed always-valid release wrapper procedure is included in Algorithm~\ref{alg:safety_wrapper}.

\begin{algorithm}[t]
\caption{Always-valid release wrapper}
\label{alg:safety_wrapper}
\begin{algorithmic}[1]
\Require Safety level \(\alpha\), maximum horizon \(T_{\max}\), fixed reference pool \(\mathcal R\), betting functions \((f_t)\)
\State Initialize \(E \gets 1\)
\For{\(t=1,2,\ldots,T_{\max}\)}
    \State Observe candidate \(C_t\) and verifier score \(S_t\)
    \State Compute
    \[
    p_t = \frac{1+\sum_{i=1}^n \mathbf 1\{R_i \ge S_t\}}{n+1}
    \]
    \State Update evidence \(E \gets E\cdot f_t(p_t)\)
    \If{\(E \ge 1/\alpha\)}
        \State \Return release \(C_t\)
    \EndIf
\EndFor
\State \Return abstain
\end{algorithmic}
\end{algorithm}

\begin{proposition}[Always-valid false-release control]
\label{prop:always_validity}
Let \(D\) be defined as above from an e-process \((E_t)_{t\ge0}\) satisfying Proposition~\ref{prop:betting_martingale}. Then
\[
\mathbb P(D=1 \mid F=0) \le \alpha.
\]
\end{proposition}

Proposition~\ref{prop:always_validity} is the primitive safety guarantee delivered by the wrapper. Combined with Proposition~\ref{prop:rout_from_primitive}, it explains why conservative false-release control on infeasible instances yields meaningful protection for the release-conditional risk of the deployed system.

\subsection{Feasible-side release}
\label{subsec:feasible_power}

The false-release guarantee in Proposition~\ref{prop:always_validity} controls the Type~I side of the release problem. A useful wrapper must also remain active on feasible instances. If safety alone were the objective, the trivial rule that always abstains would satisfy Proposition~\ref{prop:always_validity}, but it would have no deployment value because it sets the feasible-side release probability to zero. We therefore complement the safety result with an analysis of when the proposed wrapper releases under \(F=1\).

The feasible-side analysis cannot be assumption-free. When feasible and infeasible transcripts are indistinguishable to the wrapper, no black-box rule can release frequently under \(F=1\) while controlling release under \(F=0\). Proposition~\ref{prop:observable_separation} formalizes this limitation.

\begin{proposition}[Observable-separation limit for feasible-side release]
\label{prop:observable_separation}
Fix a horizon \(T\). Let \(P_f^T=\mathcal L((\mathcal R,H_T)\mid F=f)\), \(f\in\{0,1\}\), be the conditional law of the observed transcript available to the release rule by time \(T\). Then any terminal release rule \(D_T\in\{0,1\}\) that is measurable with respect to \(\sigma(\mathcal R,H_T)\) satisfies
\[
\mathbb P(D_T=1\mid F=1) \le \mathbb P(D_T=1\mid F=0)+\TV(P_1^T,P_0^T).
\]
Here \(\TV\) denotes total variation distance.
\end{proposition}

Proposition~\ref{prop:observable_separation} shows that feasible-side activity is impossible without observable separation. If a rule controls false release at level \(\alpha\), then Proposition~\ref{prop:observable_separation} gives \(\mathbb P(D_T=1\mid F=1)\le \alpha+\TV(P_1^T,P_0^T)\). Thus any method that also achieves feasible-side release probability at least \(1-\beta\) must have \(\TV(P_1^T,P_0^T)\ge 1-\beta-\alpha\). This limitation is exactly why a power statement must impose some condition on the evidence supplied by feasible trajectories.

We therefore characterize feasible-side activity through the calibrated gain supplied by correct candidates. For this, fix a betting function in Algorithm~\ref{alg:safety_wrapper} to a non-increasing \(f:[0,1]\to(0,\infty)\) satisfying \(\int_0^1 f(u)\,du\le1\). Define
% \begin{align*}
% Z_t
% &:= \log f(p_t)-\log f(1),\\
% Z_{\max}
% &:= \log f\!\left(\frac{1}{n+1}\right)-\log f(1),\\
% A_{\alpha,T}
% &:= \log(1/\alpha)-T\log f(1).
% \end{align*}
\begin{align*}
Z_t := \log f(p_t)-\log f(1) \textrm{~and~}
Z_{\max}:= \log f\!\left(\frac{1}{n+1}\right)-\log f(1).
\end{align*}
The quantity \(Z_t\) is the step-\(t\) calibrated log-gain relative to the least favorable calibrated value \(p=1\). Since \(f\) is non-increasing and \(p_t\in[1/(n+1),1]\), smaller \(p_t\)-values produce larger gains, and $0\le Z_t\le Z_{\max}.$ Thus \(Z_t\) measures how much evidence the current calibrated \(p\)-value contributes beyond the baseline value \(f(1)\), while \(Z_{\max}\) is the largest possible one-step gain under the finite reference pool. Then the threshold \(A_{\alpha,T} := \log(1/\alpha)-T\log f(1)\) is the cumulative calibrated gain needed for the terminal event \(E_T\ge1/\alpha\). Indeed, under the fixed calibrator \(f\), $E_T=\prod_{t=1}^T f(p_t),$
and hence
%\begin{align*}
$\log E_T
=\sum_{t=1}^T \log f(p_t)=\sum_{t=1}^T \{\log f(1)+Z_t\}=T\log f(1)+\sum_{t=1}^T Z_t.$
%\end{align*}
Therefore, $E_T\ge \frac{1}{\alpha}$ if and only if $\sum_{t=1}^T Z_t\ge A_{\alpha,T}.$ In words, \(A_{\alpha,T}\) is the amount of cumulative calibrated gain that must be accumulated by time \(T\) for the terminal evidence to cross the release threshold.

\begin{assumption}[Correct-candidate calibrated gain]
\label{ass:certifiable_gain}
For a fixed horizon \(T\), assume that under \(F=1\), there exists a deterministic constant \(B_T\ge0\) such that
\[
\sum_{t=1}^T
\mathbb E\!\left[
Y_t Z_t
\,\middle|\,
\widetilde{\mathcal G}_{t-1},F=1
\right]
\ge
B_T
\qquad\text{almost surely}.
\]
\end{assumption}

Assumption~\ref{ass:certifiable_gain} says that, on feasible tasks, the workflow supplies enough \emph{correct-candidate calibrated gain} over the first \(T\) steps. The factor \(Y_t\in\{0,1\}\) ensures that only gain contributed by actually correct candidates is counted: when \(Y_t=1\), the candidate is correct and \(Y_tZ_t=Z_t\); when \(Y_t=0\), the candidate is incorrect and \(Y_tZ_t=0\). Thus the assumption does not merely require favorable verifier scores in general. Rather, it requires that genuinely correct candidates contribute enough calibrated evidence, relative to the hard-negative reference pool, for the release threshold to be reachable within horizon \(T\).

This requirement is natural in applications such as coding agents. On a feasible coding task, once the workflow begins to generate correct or nearly correct programs, these candidates should tend to receive verifier scores that remain favorable relative to previously observed high-scoring failures. Even if no single step is decisive on its own, repeated appearances of such correct candidates can accumulate enough calibrated gain to support release. In this sense, Assumption~\ref{ass:certifiable_gain} formalizes the idea that feasible trajectories should repeatedly produce not just high scores, but \emph{certifiably informative} scores.

A useful interpretation is obtained by writing
\[
\pi_t := \mathbb P(Y_t=1\mid\widetilde{\mathcal G}_{t-1},F=1), 
\qquad
\bar Z_t := \mathbb E(Z_t\mid\widetilde{\mathcal G}_{t-1},Y_t=1,F=1),
\]
with the convention that \(\pi_t\bar Z_t=0\) when \(\pi_t=0\). Then
\[
\mathbb E\!\left(Y_tZ_t\mid\widetilde{\mathcal G}_{t-1},F=1\right)=\pi_t\bar Z_t.
\]
Here \(\pi_t\) is the conditional probability that the workflow reaches a correct candidate, while \(\bar Z_t\) is the calibrated gain contributed by a correct candidate when one is reached. The first term reflects generation quality; the second reflects how strongly the evaluator and reference-pool calibration distinguish correct candidates from high-scoring failures. Accordingly, Assumption~\ref{ass:certifiable_gain} requires the cumulative product of these two effects to be large enough.

\begin{theorem}[Finite-horizon release power]
\label{thm:feasible_power_cert_gain}
Assume Assumption~\ref{ass:certifiable_gain}. Let \(\Delta_{\alpha,T}=B_T-A_{\alpha,T}\). If \(\Delta_{\alpha,T}>0\) and \(Z_{\max}>0\), then
\[
\mathbb P(\tau_\alpha\le T\mid F=1)
\ge
1-
\exp\!\left\{
-\frac{2\Delta_{\alpha,T}^2}{T Z_{\max}^2}
\right\}.
\]
\end{theorem}

Theorem~\ref{thm:feasible_power_cert_gain} compares the supplied correct-candidate gain \(B_T\) with the required gain \(A_{\alpha,T}\). If \(B_T\) exceeds \(A_{\alpha,T}\), then feasible trajectories provide enough expected calibrated evidence to clear the release threshold by time \(T\), up to a martingale concentration error. The margin \(\Delta_{\alpha,T}=B_T-A_{\alpha,T}\) controls the strength of this conclusion. Larger margins give a smaller exponential error term and hence a larger lower bound on \(\mathbb P(\tau_\alpha\le T\mid F=1)\). This is the sense in which the wrapper differs from an always-abstain rule. Under feasible trajectories that repeatedly supply positive calibrated gain from correct candidates, the same conservative rule can still cross the release threshold within the interaction horizon.

%The result should be read as a feasible-side activity statement. Proposition~\ref{prop:always_validity} controls false release under \(F=0\), while Theorem~\ref{thm:feasible_power_cert_gain} identifies conditions under which the same conservative wrapper need not behave like an always-abstain rule under \(F=1\). The theorem does not assert that every released candidate is correct. Rather, it lower bounds the probability of crossing the release threshold on feasible tasks when correct candidates repeatedly provide enough calibrated evidence. The bound is conservative because the proof lower bounds the hitting event \(\{\tau_\alpha\le T\}\) by the sufficient terminal event \(\{E_T\ge1/\alpha\}\).

%This is the operational source of feasible-side release. A workflow may fail to release because it rarely reaches correct candidates, because correct candidates are not sufficiently separated from hard negatives after calibration, or because the interaction horizon is too short for moderate gains to accumulate.

\begin{remark}[Generator and verifier improvements]
Theorem~\ref{thm:feasible_power_cert_gain} also describes how the wrapper should benefit as the underlying AI workflow improves. A stronger generator helps when it increases \(\pi_t\), the chance that the workflow reaches a correct candidate. A stronger verifier helps when it increases \(\bar Z_t\), the calibrated gain assigned to correct candidates relative to the reference pool. Thus future improvements in AI translate into better release behavior only insofar as they increase the correct-candidate gain \(B_T\). For fixed \(f\), \(n\), and \(T\), if an improved generator-verifier pair raises this lower bound from \(B_T\) to \(B_T'\ge B_T+\delta\), then the release margin increases from \(\Delta_{\alpha,T}\) to at least \(\Delta_{\alpha,T}+\delta\). Overall, what matters is that correct candidates become more frequent, more separable from high-scoring failures, or both.
\end{remark}

\subsection{Calibration drift}
\label{subsec:dynamic_drift}

The always-valid false-release control in Proposition~\ref{prop:always_validity} assumes exact failure-side calibration dominance. In deployment, however, repeated self-conditioning and other forms of history dependence may gradually shift the online score law away from the nominal calibration target. This subsection quantifies how such stepwise misspecification affects the false-release guarantee over a finite interaction horizon. 
% Proposition~\ref{prop:always_validity} is stated under exact failure-side calibration dominance. In deployment, repeated self-conditioning and other forms of history dependence may slowly move the online score law away from the nominal calibration target. This subsection quantifies how such stepwise misspecification inflates the false-release bound over a finite interaction horizon. It should not be read as a literal model of unbounded reasoning. Practical horizons are constrained by token budgets, latency, compute cost, and engineering timeouts.
Let \(P_0\) denote a nominal reference law for the stepwise \(p\)-value stream, and under \(F=0\) let \(Q_{t,0}=\mathcal L(p_t\mid\widetilde{\mathcal G}_{t-1},F=0)\).

\begin{theorem}[Robustness to calibration drift]
\label{thm:drift_robustness}
Let \(e_t=f(p_t)\in[0,M]\), \(E_t=\prod_{j=1}^t e_j\), and \(\tau_\alpha=\inf\{t\ge1:E_t\ge1/\alpha\}\), where \(\int_0^1 f(u)\,P_0(du)\le1\). Assume that, under \(F=0\), \(\TV(Q_{t,0},P_0)\le\epsilon\) almost surely for all \(t\ge1\). Then, for any \(T\in\mathbb N\),
\[
\mathbb P\!\left(\tau_\alpha\le T \,\middle|\, \sigma(\mathcal R),F=0\right)
\le
\alpha(1+M\epsilon)^T.
\]
\end{theorem}

Theorem~\ref{thm:drift_robustness} gives a finite-horizon view of calibration drift. In the normalized case \(M=1\), the inflation factor is \((1+\epsilon)^T\), which is close to \(\exp(T\epsilon)\) for small \(\epsilon\). The relevant quantity is therefore the accumulated mismatch \(T\epsilon\), not the one-step mismatch alone. This is important in AI workflows because \(T\) is the length of an interaction trajectory rather than an asymptotic sample size. In practice, this horizon is limited by token, latency, compute budgets, and engineering timeouts, which are already treated as operational constraints in coding-agent systems \citep{anthropic2026_claude_code_costs,anthropic2026_claude_code_monitoring}. The result therefore quantifies how finite-horizon safety degrades under small calibration drift, while making clear that such drift cannot be ignored over arbitrarily long interaction horizons.
}

\section{Case Study: MBPP+ Code Generation}
\label{sec:mbpp_case_study}

We study the proposed wrapper on MBPP+ (Mostly Basic Python Problems Plus), a code-generation benchmark \citep{austin2021program,liu2023your}. The benchmark provides a coding-agent instance of the broader workflow release problem: during deployment, the wrapper observes only a lightweight visible-test verifier, while offline correctness is assessed using a larger hidden test suite. This visible-hidden mismatch makes premature release a practical risk, since a generated program can receive a favorable visible score while still failing hidden tests. The case study therefore asks whether an adaptive generator-verifier workflow can accumulate enough calibrated evidence to release without relying on raw verifier scores alone.

Our implementation uses \texttt{Qwen2.5-Coder-7B-Instruct} \citep{hui2024qwen2} with \texttt{max\_new\_tokens}=240, \texttt{temperature}=0.7, and \texttt{top\_p}=0.95. For each task, the model generates a trajectory of ten sequential candidates, with one candidate produced at each step. Throughout the main analysis, the online verifier score \(S_t\in[0,1]\) is the empirical pass rate on a visible subset of unit tests. The wrapper observes only this visible score stream and must decide whether to release or abstain. Sections~\ref{app:implementation} and~\ref{sec:supp_numericals} of the Supplementary Material provide implementation details and additional numerical results.

\subsection{Task curation, visible-hidden split, and empirical regimes}
\label{subsec:mbpp_setup}

Each MBPP+ task is a Python coding problem consisting of four components:
\begin{enumerate}[leftmargin=1.5em, nosep]
    \item \textbf{Prompt:} A natural-language instruction describing the desired logic.
    \item \textbf{Entry point:} The exact function name the LLM is required to implement.
    \item \textbf{Canonical solution:} A human-verified reference implementation.
    \item \textbf{Test cases:} A collection of input-output pairs used to evaluate generated code.
\end{enumerate}
{
In our study, the test cases are partitioned into two disjoint subsets. The first serves as a \emph{visible verifier}, providing a lightweight online signal to score candidates during deployment. The second acts as a stronger \emph{hidden oracle} for offline evaluation. This visible-hidden split creates the central practical challenge: a generated program may achieve a highly favorable visible score yet fail the hidden tests, rendering the online verifier informative but imperfect.
}

The following representative task illustrates this mismatch. It also helps explain why selective release is nontrivial in code generation. 

\textbf{Representative MBPP+ task: structural pattern matching.}
Task \texttt{Mbpp/74} asks whether two sequences follow the same structural pattern. This type of task is vulnerable to verifier mismatch because a limited visible test subset may capture only basic positive or negative matches while missing important edge cases such as length mismatches, repeated sub-patterns, or empty-list behavior. As a result, a structurally flawed candidate can overfit the visible verifier and obtain a deceptively high score while still failing the hidden oracle.

\begin{tcolorbox}[sharp corners, boxrule=0.5pt, colback=gray!5, fontupper=\small]
\textbf{Task \texttt{Mbpp/74}}\\
\textbf{Entry point:} \texttt{is\_samepatterns}\\[2pt]
\footnotesize
\textbf{Prompt:}
\begin{verbatim}
Write a function to check whether the words in a list follow the sequence 
given in the patterns array.
assert is_samepatterns(["red", "green", "green"], ["a", "b", "b"]) == True
\end{verbatim}
\textbf{Canonical solution:}
\begin{verbatim}
def is_samepatterns(words, patterns):
    if len(words) != len(patterns):
        return False
    return len(set(words)) == len(set(patterns)) == len(set(zip(words, patterns)))
\end{verbatim}
\textbf{Example Verifier inputs:}
\begin{itemize}[noitemsep, leftmargin=1.5em]
\item \texttt{[[ ], [ ]]}  (Expected: \texttt{True})
\item \texttt{[["red", "green", "red"], ["a", "b", "c", "a", "a"]]} (Expected: \texttt{False})
\end{itemize}
\textbf{Example Hidden Oracle inputs:}
\begin{itemize}[noitemsep, leftmargin=1.5em]
\item \texttt{[["a", "a", "b", "bblueccc", "b", "b"], ["a", "a", "b", "bblueccc", "b", "b"]]} (Expected: \texttt{True})
\item \texttt{[["reed", "blue", "red", "red", "red"], ["reed", "blue", "red", "red", "red"]]} (Expected: \texttt{True})
\end{itemize}
\end{tcolorbox}

To standardize evaluation, we retain only tasks with at least \(100\) usable tests, where a usable test is one passed by the canonical solution. For each retained task, we deterministically select \(100\) tests. These are partitioned into \(30\) visible verifier tests and \(70\) hidden evaluation tests. The visible tests are used to compute the online verifier score, while the hidden tests are used only for offline evaluation and empirical feasibility labeling. The main paper focuses on the \(N_{\mathrm{test}}=30\) setting; reduced verifier sizes \(N_{\mathrm{test}}\in\{10,20\}\) are discussed in Supplementary Section~\ref{subsec:mbpp_robustness} for robustness analysis.

We further partition the curated tasks into three disjoint subsets. The \emph{bank split} is used only for constructing candidate reference pools and calibrating the entropy baseline. The \emph{select split} is used only for diagnosing candidate reference-pool families. The \emph{final split} is reserved exclusively for held-out reporting. This separation is important: the final evaluation is not used to choose the pool family, and the select split is not used for final performance reporting. In the present study, the selected family is \texttt{top55}, which retains the upper \(55\%\) of executable incorrect bank-step scores, with ties included at the quantile cutoff.

Because true algorithmic correctness is not available in an absolute sense, we define empirical correctness by whether a generated candidate passes all \(70\) hidden tests. Based on this hidden-test evaluation over the full ten-step trajectory, we partition the held-out tasks into an empirical infeasible subset \(F_0\) and an empirical feasible subset \(F_1\). The subset \(F_0\) contains tasks for which no correct candidate ever appears, while \(F_1\) contains tasks for which at least one correct candidate appears. On \(F_0\), every release is necessarily an error, so the release rate coincides with the false-release rate. On \(F_1\), the release rate serves as a power-style companion metric, while the release-conditional failure rate measures the reliability of released outputs.

\subsection{Baselines}
\label{subsec:mbpp_baselines}

We compare the proposed wrapper with three baselines motivated by existing stopping heuristics. More details of these baselines are deferred to the Supplementary Section~\ref{app:implementation}.

\begin{itemize}[leftmargin=1.5em]
    \item \textbf{\texttt{First\_p}.} A single-step calibrated stopping rule, defined by
    \[
    \tau=\inf\{t:p_t\le \alpha\}.
    \]
    This baseline uses the same calibrated \(p\)-values and the same reference pool as our method, but without sequential evidence accumulation.
{
    \item \textbf{\texttt{Entropy}.} A predictive-confidence baseline based on entropy \citep{sharma2025think}: It releases once the candidate's mean entropy falls below a threshold estimated from correctly generated bank-split candidates.

    \item \textbf{\texttt{Stability}.} A short-run score-convergence baseline motivated by recent stopping heuristics based on response stability \citep{li2024escape, mao2025early}. It releases when the visible verifier score becomes sufficiently stable at a sufficiently high level.}
\end{itemize}
The \texttt{Entropy} and \texttt{Stability} baselines do not depend on \(\alpha\) or on the reference pool. The \texttt{First\_p} baseline isolates the role of sequential accumulation by using the same calibrated \(p\)-values as our method but making a release decision from a single step rather than from accumulated evidence. In our experiments, the method in \cite{xie2026statistical} is not applicable because their method relies on token-level uncertainty signals from reasoning traces and targets early stopping for well-posed queries, whereas our framework is designed as a black-box wrapper for false-release control.

A useful point of reference is the degenerate rule that always abstains. Such a rule trivially guarantees zero false releases, but it also yields a feasible-side release rate of zero and is therefore uninformative in practice. Accordingly, our evaluation tracks not only false-release control on \(F_0\), but also feasible-side activity on \(F_1\), so that conservative behavior is not confused with useful deployment performance.

%{\color{blue}Before evaluating the empirical performance, consider a trivial baseline: a wrapper that always abstains from release. Such a rule guarantees zero false releases, but it also yields a feasible-side release rate of zero and is therefore uninformative in practice. For this reason, the relevant question is not only whether a method controls false release under \(F_0\), but also whether it preserves meaningful feasible-side activity under \(F_1\). Our empirical evaluation therefore tracks both quantities.}\Young{moved the old ``trivial abstention'' remark to here.}

\subsection{Reference-pool construction and held-out selection}
\label{subsec:mbpp_pool_selection}

Following Section~\ref{subsec:reference_pool}, we construct candidate reference pools from incorrect trajectories in the bank split, retaining only the upper tail of the visible verifier score. A broad pool such as \texttt{all\_incorrect}, which includes every incorrect candidate regardless of its score, dilutes the failure-side upper tail with easy negatives and makes online scores appear artificially extreme. At the opposite extreme, overly restrictive pools such as \texttt{top35} or \texttt{top45} may become too conservative and suppress feasible-side release within the ten-step horizon. We therefore use the select split to choose a practical compromise. 

\begin{figure}[t]
\centering
\includegraphics[width=0.7\textwidth]{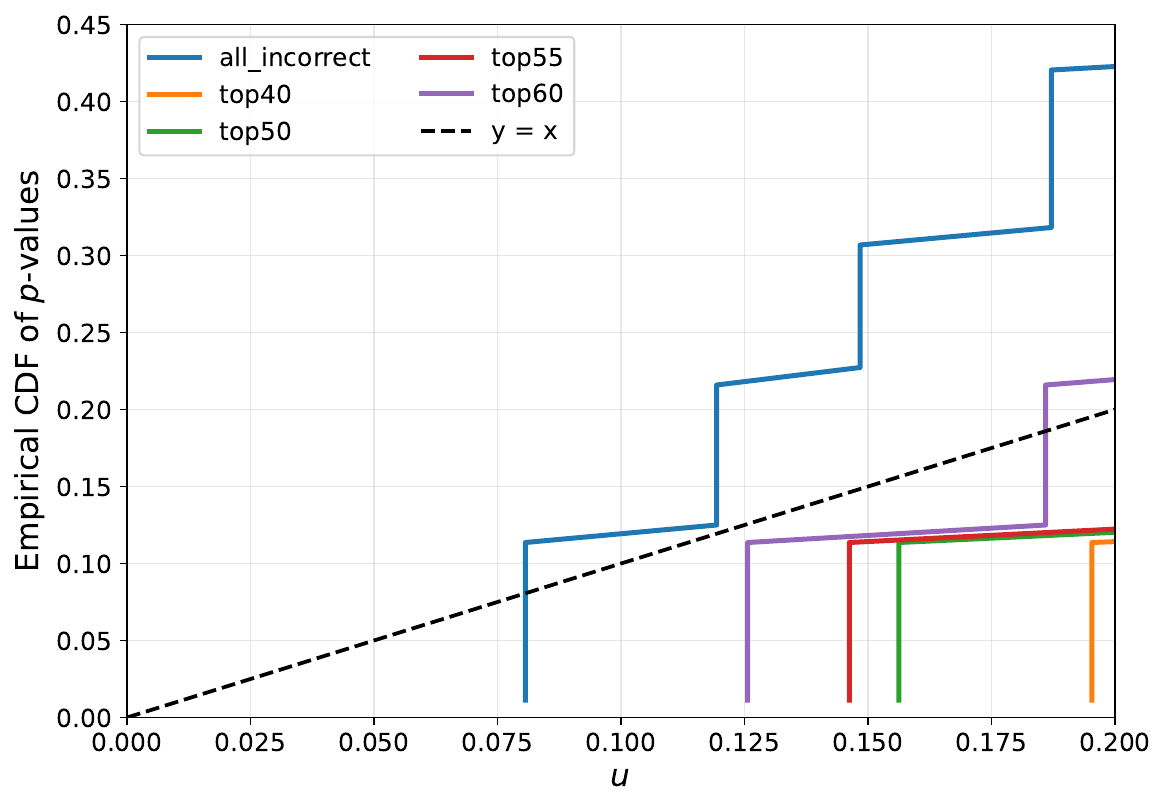}
\caption{\textbf{Held-out reference-pool diagnostic at \(N_{\mathrm{test}}=30\).} The broad pool \texttt{all\_incorrect} is anti-conservative, while increasingly aggressive upper-tail pools become more conservative. The selected \texttt{top55} pool is the most relaxed retained pool that passes the held-out diagnostic.}
\label{fig:main_diag_top55}
\end{figure}

Figure~\ref{fig:main_diag_top55} reports the select-stage diagnostic at \(N_{\mathrm{test}}=30\). The broad pool \texttt{all\_incorrect} is visibly anti-conservative. As the retained pool is restricted to harder negatives, the induced empirical upper-tail \(p\)-values move below the diagonal, indicating increasingly conservative calibration. Among the candidate families considered in the main paper, \texttt{top55} is the most relaxed pool that passes the held-out diagnostic. Stricter pools do not yield an obvious additional safety benefit in this diagnostic, while relaxing the threshold further produces visible lower-tail inflation. We therefore use \texttt{top55} in all subsequent main-paper evaluations. Supplement Section~\ref{subsec:pool_family_summary} reports a broader sensitivity analysis over candidate upper-tail families. It confirms that \texttt{top55} is the least conservative choice among those considered that still passes the held-out diagnostic.

%A broader scan over candidate pool families is reported in the supplement. 

%{\color{blue} A broader sensitivity analysis of candidate upper-tail families is reported in Supplement Section~\ref{subsec:pool_family_summary}. It confirms that \texttt{top55} is the most relaxed pool among the candidates considered that still passes the held-out conservativeness diagnostic. There we compare a wider range of reference-pool constructions across verifier settings, whereas the main text focuses on the selected \texttt{top55} pool and its held-out implications for release behavior.}

\subsection{Evaluation of false-release control and feasible-side release}
\label{subsec:mbpp_main_results}

Table~\ref{tab:mbpp_main_top55} reports the held-out results at \(N_{\mathrm{test}}=30\) using the selected \texttt{top55} reference pool. On the empirical infeasible subset \(F_0\), the primary metric is the false-release rate. On the empirical feasible subset \(F_1\), we report the release rate, the release-conditional failure rate, and the average release step. 

The heuristic baselines \texttt{Entropy} and \texttt{Stability} exhibit elevated false-release rates under \(F_0\). The entropy rule relies on point-in-time predictive confidence, while the stability rule relies on short-run smoothness of the visible pass rate. In this setting, neither local confidence nor local smoothness is sufficient to protect against spuriously favorable verifier signals from structurally incorrect candidates. Their false-release rates on \(F_0\) are \(0.7667\) and \(0.2333\), respectively. The \texttt{First\_p} baseline isolates the role of the decision layer, since it applies a one-step threshold to the exact same calibrated $p$-values used by our e-process. Under the selected hard-negative pool, the induced $p$-value scale is discrete. In the main setting, because the selected pool of 170 hard negatives already contains 24 incorrect bank steps with perfect visible scores, even a flawless 30/30 maps only to $p_t=(1+24)/(170+1) \approx 0.146$. Consequently, at the more relaxed threshold $\alpha=0.20$, \texttt{First\_p} is highly active: it releases on 99.32\% of feasible tasks, but this aggressiveness comes with a false-release rate of 0.0667 on $F_0$. At stricter levels, however, it encounters a sharp utility cliff. When $\alpha \in \{0.10, 0.05\}$, no single calibrated step is sufficient for release, and the method becomes inactive on feasible held-out tasks. In this benchmark, the limitation of a one-step calibrated rule is therefore not only its vulnerability to spurious early signals under $F_0$, but also the resolution limit it faces under $F_1$.

By contrast, the proposed wrapper overcomes this limit by accumulating repeated moderate evidence across the trajectory. At $\alpha=0.10$, it yields zero false releases on $F_0$ and zero release-conditional failures on $F_1$, while maintaining a feasible-side release rate of 0.7703. At $\alpha=0.05$, it remains active on 74.32\% of feasible tasks, with a release-conditional failure rate of 0.0091. This improvement comes at the cost of later release: because the e-process requires sustained favorable evidence to cross the threshold, the average release step is naturally delayed relative to more aggressive one-step rules. Overall, the comparison shows that sequential accumulation preserves useful feasible-side activity precisely where one-step calibrated stopping breaks down.

{\color{black}
This comparison also illustrates the role of Theorem~\ref{thm:feasible_power_cert_gain}. False-release control alone would be achieved by an always-abstain rule, but such a rule has no feasible-side activity. The proposed wrapper is different: at \(\alpha=0.10\), it has zero false release on \(F_0\) while still releasing on \(77.03\%\) of \(F_1\) tasks. The empirical mechanism is consistent with the gain-based interpretation in Theorem~\ref{thm:feasible_power_cert_gain}. Release does not require a single decisive calibrated step; repeated favorable candidates can supply enough cumulative calibrated gain to cross the evidence threshold.
}

\begin{table}[t]
\centering
\caption{MBPP+ held-out results at \(N_{\mathrm{test}}=30\) using the selected \texttt{top55} reference pool. On \(F_0\), the false-release rate is the primary safety metric. On \(F_1\), we report the release rate, the release-conditional failure rate, and the average release step.}
\label{tab:mbpp_main_top55}
\small
\begin{tabular}{llcccccc}
\toprule
&& \multicolumn{2}{c}{$F_0$ $(n=30)$} && \multicolumn{3}{c}{$F_1$ $(n=148)$} \\
\cmidrule{3-4} \cmidrule{6-8}
Method & $\alpha$ & False-release & Release step && Release rate & Failure$\,|$release & Release step \\
\midrule
\texttt{First\_p}  & 0.20 & 0.0667 & 3.5000 && 0.9932 & 0.0272 & 1.4150 \\
\texttt{First\_p}  & 0.10 & 0.0000 & ---    && 0.0000 & ---    & --- \\
\texttt{First\_p}  & 0.05 & 0.0000 & ---    && 0.0000 & ---    & --- \\
\midrule
Ours      & 0.20 & 0.0333 & 10.0000 && 0.8041 & 0.0000 & 4.3025 \\
Ours      & 0.10 & 0.0000 & ---     && 0.7703 & 0.0000 & 6.2719 \\
Ours      & 0.05 & 0.0000 & ---     && 0.7432 & 0.0091 & 7.2182 \\
\midrule
\texttt{Entropy}   & ---  & 0.7667 & 1.7826 && 0.7838 & 0.1207 & 2.0345 \\
\texttt{Stability} & ---  & 0.2333 & 2.7143 && 0.9324 & 0.0290 & 2.4565 \\
\bottomrule
\end{tabular}
\end{table}

\subsection{Representative task trajectories}
\label{subsec:mbpp_traces}

Aggregate metrics alone do not fully reveal how the wrapper behaves along individual reasoning trajectories. We therefore examine two representative held-out tasks at \(\alpha=0.10\), chosen to illustrate the two main operating modes of the method: safe abstention under persistent verifier deception and delayed correct release through evidence accumulation. %Full task specifications are provided in the supplement.

\begin{table}[!t]
\centering
\caption{Representative trajectories. In \texttt{Mbpp/74}, the wrapper safely abstains despite persistently high verifier scores on an incorrect trajectory, whereas \texttt{Entropy} and \texttt{Stability} both release incorrectly. In \texttt{Mbpp/598}, \texttt{Entropy} releases incorrectly at \(t=1\), \texttt{Stability} releases correctly at \(t=3\), \texttt{First\_p} never releases, and our wrapper accumulates evidence and eventually releases the correct solution.}
\label{tab:mbpp_task_traces}
\small
\resizebox{\textwidth}{!}{
\begin{tabular}{lccccc}
\toprule
Step & Score & Correct & \(p_t\) & Wealth \(E_t\) & Outcome \\
\midrule
\multicolumn{6}{l}{\textit{\textbf{Scenario A: Safe abstention against persistent deception (\texttt{Mbpp/74})}}} \\
1  & 0.967 & False & 0.216 & 1.185 & \texttt{Entropy}: release (error) \\
2  & 0.967 & False & 0.216 & 1.405 & \texttt{Stability}: release (error) \\
$\vdots$ & $\vdots$ & $\vdots$ & $\vdots$ & $\vdots$ & $\vdots$ \\
10 & 0.967 & False & 0.216 & 5.469 & Ours/\texttt{First\_p}: no release \\
\midrule
\multicolumn{6}{l}{\textit{\textbf{Scenario B: Delayed correct release via evidence accumulation (\texttt{Mbpp/598})}}} \\
1  & 0.867 & False & 0.298 & 0.947  & \texttt{Entropy}: release (error) \\
2  & 1.000 & True  & 0.146 & 1.476  & --- \\
3  & 1.000 & True  & 0.146 & 2.302  & \texttt{Stability}: release (correct) \\
$\vdots$ & $\vdots$ & $\vdots$ & $\vdots$ & $\vdots$ & $\vdots$ \\
7  & 1.000 & True  & 0.146 & 13.617 & Ours: release (success) \\
$\vdots$ & $\vdots$ & $\vdots$ & $\vdots$ & $\vdots$ & $\vdots$ \\
10 & 1.000 & True  & 0.146 & 51.644 & \texttt{First\_p}: no release \\
\bottomrule
\end{tabular}
}
\end{table}

\textbf{Safe abstention under persistent verifier deception.}
Task \texttt{Mbpp/74} asks whether two sequences follow the same pattern structure. In this type of structural matching problem, a limited visible test subset may miss failures involving length, alignment, or repeated-subpattern logic. In the observed trajectory, no correct candidate ever appears under the hidden oracle, yet the visible verifier score remains \(0.967\) throughout. Both \texttt{Entropy} and \texttt{Stability} are triggered by these persistently favorable local signals and release incorrectly. Under the selected hard-negative pool, however, the score calibrates to \(p_t=0.216\), which is not especially extreme relative to previously observed deceptive failures. As a result, the accumulated evidence grows only gradually, reaching \(5.469\) by the terminal step and never crossing the release threshold of \(10\). In this case, the wrapper avoids false release because it evaluates the score relative to a benchmark of strong failures rather than relying on its raw magnitude alone.

{\textbf{Correct release through repeated moderate evidence.} The task \texttt{Mbpp/598} asks whether an integer is an Armstrong number. Such arithmetic predicate tasks are prone to partial overfitting: a flawed rule may perform well on the visible verifier subset while failing under the hidden oracle. At \(t=1\), an incorrect candidate receives a visible score of \(0.867\), which prompts \texttt{Entropy} to release incorrectly. From \(t=2\) onward, the generated logic is correct and the visible score is \(1.000\). Under the selected hard-negative pool, however, even a perfect visible score maps only to \(p_t=0.146\). While this is sufficient for \texttt{Stability} to release at \(t=3\), it remains strictly above the \(\alpha=0.10\) threshold for \texttt{First\_p}, which therefore never releases despite the task being effectively solved from \(t=2\) onward. The proposed wrapper succeeds by accumulating this repeated moderate evidence over time. As the correct candidate continues to appear, the wealth process grows from \(1.476\) to \(13.617\), crossing the release threshold at \(t=7\).
In the notation of Theorem~\ref{thm:feasible_power_cert_gain}, this is a trajectory where no single calibrated step is decisive, but the repeated correct candidates after \(t=2\) contribute positive calibrated gains \(Z_t\). The release at \(t=7\) is therefore a cumulative-gain crossing rather than a one-step threshold crossing.
 %This trajectory illustrates that the gain from sequential accumulation is not that it produces a more favorable calibrated step, but that it converts repeatedly favorable yet individually insufficient evidence into a valid release.

Overall, the MBPP+ case study illustrates the two main operating characteristics of the wrapper. On empirically infeasible trajectories, it resists repeated but non-decisive verifier spikes and avoids premature release. On empirically feasible trajectories, it remains active by accumulating moderate evidence across steps until the evidence threshold is crossed. The examples show how the wrapper can abstain under persistent verifier deception while still releasing when useful calibrated evidence appears repeatedly over the trajectory.
}

\section{Conclusion}

We studied release decisions for black-box generate-verify AI workflows, where the central question is when evidence accumulated along an adaptive trajectory is strong enough to justify releasing an output. We formulated this as a sequential selective-release problem. The proposed wrapper separates score calibration from evidence accumulation. A hard-negative reference pool converts adaptive verifier scores into calibrated evidence, and an e-process accumulates that evidence under optional stopping. This yields finite-sample false-release control under a failure-side dominance condition and explains how the same conservative rule can remain active on feasible instances through repeated calibrated gain. In the MBPP+ case study, the wrapper reduced premature incorrect release relative to heuristic stopping rules while preserving feasible-side release through evidence accumulation.

More broadly, AI-assisted decision-making is creating new statistical objects beyond fixed datasets and one-shot predictions. This paper studies one such setting, but many others are likely to arise. A central opportunity for statistics would be to develop principled decision rules for these new structures, especially when classical assumptions are unavailable.

% \section*{Acknowledgements}

% The authors used ChatGPT 5.5 to assist with language polishing, organization, and editorial refinement. The authors also used OpenAI Codex to assist with code polishing, implementation organization, and reproducibility checks for the numerical study. The research ideas, methodological development, theoretical results, empirical design, and final manuscript decisions are the authors' own.

\spacingset{1.8} % DON'T change the spacing!
\bibliography{bibliography}

\newpage
\appendix
\setcounter{section}{0}
\setcounter{table}{0}
\setcounter{figure}{0}
\setcounter{equation}{0}
\renewcommand{\thesection}{\Alph{section}}
\renewcommand{\thetable}{S\arabic{table}}
\renewcommand{\thefigure}{S\arabic{figure}}
\renewcommand{\theequation}{S\arabic{equation}}

\spacingset{1.3}
\begin{center}    
{\large\bf SUPPLEMENTARY MATERIAL of \\
``When Should an AI Workflow Release? \\ Always-Valid Inference for Black-Box Generate-Verify Systems''}
\end{center}

\bigskip

\noindent
This supplementary material provides the implementation details, auxiliary diagnostics, and proofs that support the main text. %The goal is to keep the main paper focused on the release problem, the statistical wrapper, and the primary MBPP+ evidence, while collecting benchmark-specific construction details and extended diagnostic checks here.

\begin{itemize}[leftmargin=1.6em, itemsep=0.25em, topsep=0.25em]
    \item Section~\ref{app:implementation} documents the MBPP+ case-study implementation, including representative tasks, task curation, visible-hidden test construction, empirical feasible and infeasible subsets, baseline rules, and computational environment.

    \item Section~\ref{sec:supp_numericals} reports additional MBPP+ studies beyond the main text. This includes the broader reference-pool diagnosis in Section~\ref{subsec:pool_family_summary}, robustness checks in Section~\ref{subsec:mbpp_robustness}, an additional trajectory in Section~\ref{subsec:additional_traces}, and certifiable-gain diagnostics in Section~\ref{subsec:sim_theory}.

    \item Section~\ref{app:proofs} delivers the proofs of the formal results in the main paper.
\end{itemize}

\medskip

\section{Details on the MBPP+ Case Study}
\label{app:implementation}

This section records the benchmark-specific implementation details behind the MBPP+ case study. It begins with two additional representative task specifications that complement the main-text example, then describes how usable input-output pairs are curated and split into visible-verifier and hidden-oracle sets. The subsequent subsections explain the bank, select, and final task splits, the empirical construction of feasible and infeasible held-out subsets, the entropy and stability baselines, and the computational environment used to generate and evaluate the cached trajectories.

\subsection{Additional representative MBPP+ tasks}
\label{sec:supp_task_specs}

The main text introduces the MBPP+ setup and \texttt{Mbpp/74} as a representative example. Here we record the remaining task specifications used in the analysis, \texttt{Mbpp/598} and \texttt{Mbpp/643}.

\noindent\textbf{Task \texttt{Mbpp/598}: Out-of-Distribution Arithmetic}. Detecting an Armstrong number is an arithmetic predicate problem. Simple heuristic logic often passes the small integers present in the verifier subset. However, structural flaws in the model's arithmetic reasoning are aggressively exposed by the unusually large integers (e.g., 20-digit numbers).

\begin{tcolorbox}[sharp corners, boxrule=0.5pt, colback=gray!5, fontupper=\small]
\spacingset{1.1}
\textbf{Task \texttt{Mbpp/598}}\\
\textbf{Entry point:} \texttt{armstrong\_number}\\[2pt]
\footnotesize
\textbf{Prompt:}
\begin{verbatim}
Write a function to check whether a given integer is an Armstrong number or not.
An Armstrong number of three digits is an integer such that the sum of the cubes 
of its digits is equal to the number itself.
assert armstrong_number(153) == True
\end{verbatim}
\textbf{Canonical solution:}
\begin{verbatim}
def armstrong_number(number):
    num_str = str(number)
    num_len = len(num_str)
    return number == sum(int(digit) ** num_len for digit in num_str)
\end{verbatim}
\textbf{Example Verifier inputs:}
\begin{itemize}[noitemsep, leftmargin=1.5em]
\item \texttt{[46]} (Expected: \texttt{False})
\item \texttt{[12345678901234567882]} (Expected: \texttt{False})
\end{itemize}
\textbf{Example Hidden Oracle inputs:}
\begin{itemize}[noitemsep, leftmargin=1.5em]
\item \texttt{[90]} (Expected: \texttt{False})
\item \texttt{[12345678901234567878]} (Expected: \texttt{False})
\end{itemize}
\end{tcolorbox}

\vspace{1em}
\noindent\textbf{Task \texttt{Mbpp/643}: Boundary Logic and Regex}. This task involves checking if `z' appears inside a word, not at the boundaries. A candidate code might capture local patterns correctly on the verifier but fail on the edge-case.

\begin{tcolorbox}[sharp corners, boxrule=0.5pt, colback=gray!5, fontupper=\small]
\spacingset{1.3}
\textbf{Task \texttt{Mbpp/643}}\\
\textbf{Entry point:} \texttt{text\_match\_wordz\_middle}\\[2pt]
\footnotesize
\textbf{Prompt:}
\begin{verbatim}
Write a python function to check if a string contains 'z', 
except at the start or end of the word.
assert text_match_wordz_middle("pythonzabc.") == True
\end{verbatim}
\textbf{Canonical solution:}
\begin{verbatim}
import re
def text_match_wordz_middle(text):
    return bool(re.search(r'\Bz\B', text))
\end{verbatim}
\textbf{Example Verifier inputs:}
\begin{itemize}[noitemsep, leftmargin=1.5em]
\item \texttt{["baaz"]} (Expected: \texttt{False})
\item \texttt{["zzzxyazzzzzzzzbczzz"]} (Expected: \texttt{True})
\end{itemize}
\textbf{Example Hidden Oracle inputs:}
\begin{itemize}[noitemsep, leftmargin=1.5em]
\item \texttt{["yyyzzz"]} (Expected: \texttt{True})
\item \texttt{["azazzbzzzczzzzzzbzzzzzzczxyabczzzzzzzxyabxczzz"]} (Expected: \texttt{True})
\end{itemize}
\end{tcolorbox}

\subsection{Task curation and pair construction}

From the MBPP+ benchmark, we retain only tasks with at least \(100\) usable input-output pairs. For each retained task, we deterministically select \(100\) pairs. This curation step standardizes the task-level evaluation budget across tasks and removes variability caused purely by heterogeneous raw pair counts. It also fixes the downstream task partition and prevents numerical comparisons from being driven by shifting task-specific evaluation sizes.

For each curated task, we reserve \(70\) oracle-label pairs and \(30\) verifier pairs. The oracle-label set is used only for empirical feasibility labeling and final correctness evaluation. The verifier score is computed from a prefix of the \(30\) verifier pairs, which yields the aligned settings \(N_{\mathrm{test}}\in\{10,20,30\}\). Throughout, the main setting is \(N_{\mathrm{test}}=30\).

\subsection{Task splits and their roles}

We partition the curated tasks into a bank split, a select split, and a final split. The bank split is used only for reference-pool construction and for entropy-threshold calibration. The select split is used only for diagnosis of candidate pool families. The final split is used only for held-out reporting. This separation is important. The diagnosis stage is not a reporting stage, and the final evaluation is not used to choose the pool family.

In the present study, the final selected family is \texttt{top55}, which retains the upper \(55\%\) of executable incorrect bank steps, with ties included at the quantile cutoff. The broad family \texttt{all\_incorrect} is used only as a diagnostic baseline.

\subsection{Empirical \(F_0\) and \(F_1\)}

For a fixed task trajectory, the empirical infeasibility subset \(F_0\) consists of tasks for which no correct candidate ever appears under the oracle-label set. The empirical feasible subset \(F_1\) consists of tasks for which at least one correct candidate appears. On \(F_0\), every release is necessarily wrong, so the release rate coincides with the false-release rate. On \(F_1\), the release rate serves as a power-style companion metric, while the release-conditional failure rate measures the reliability of released outputs.

\subsection{Detail on Baseline Methods}

We compare the proposed wrapper with two heuristic stopping rules.

\noindent\textbf{Predictive entropy.}
For candidate \(C_t\), let \(H_t\) denote the mean token-level entropy,
\[
H_t
=
\frac{1}{L_t}\sum_{j=1}^{L_t}
\left(
-\sum_{v\in\mathcal V_K}\hat\pi_{t,j}(v)\log \hat\pi_{t,j}(v)
\right),
\]
where \(\hat\pi_{t,j}\) is the predictive distribution over the top-\(K\) tokens at position \(j\), with \(K=50\) in our experiments, and \(L_t\) is the sequence length. Let \(h\) be the mean entropy over correctly generated candidates in the bank split. The entropy baseline releases at
\[
\tau_{\mathrm{ent}}:=\inf\{t\ge1:H_t\le h\},
\]
with abstention if this event does not occur within the maximum horizon.

\noindent\textbf{Algorithmic stability.}
This rule releases once the observed verifier score becomes sufficiently stable at a high level. Define
\[
\Delta_t := \max_{j\in\{t-1,t\}} S_j-\min_{j\in\{t-1,t\}} S_j.
\]
For a verifier using \(N_{\mathrm{test}}\) visible tests, the stability baseline releases at
\[
\tau_{\mathrm{stab}}
:=
\inf\{t\ge2:\Delta_t\le 1/N_{\mathrm{test}}
\text{ and } S_t\ge 0.8\},
\]
with abstention otherwise. The tolerance \(1/N_{\mathrm{test}}\) corresponds to one visible test on the discrete
verifier-score scale.

%{\color{blue}
\noindent\textbf{Computational environment.}
All computational experiments were implemented in Python, and the code-generation task was run on a single NVIDIA A100 GPU. The main computational cost lies in generating and caching the task trajectories. Once the trajectories and verifier scores are recorded, the wrapper evaluation, baseline comparisons, and certifiable-gain diagnostics are performed offline and add negligible additional cost.
%}

\section{Additional Results on MBPP+ Benchmark}
\label{sec:supp_numericals}

%{\color{blue}
This section complements the main MBPP+ analysis in four directions. Section~\ref{subsec:pool_family_summary} broadens the select-stage diagnosis over a wider range of upper-tail pool families. Section~\ref{subsec:mbpp_robustness} reports robustness checks for verifier resolution and betting aggressiveness. Section~\ref{subsec:additional_traces} gives an additional task-level trajectory illustrating oscillating feasible-side success. Section~\ref{subsec:sim_theory} outlines cached-trajectory certifiable-gain diagnostics that connect the feasible-side power analysis to the same MBPP+ trajectories used in the case study.
%}

\subsection{Broader diagnosis over candidate reference-pool families}
\label{subsec:pool_family_summary}

The main text reports the select-stage diagnosis for a small set of candidate families centered around \texttt{top55}. Table~\ref{tab:supp_diag_compact} broadens this search to \texttt{all\_incorrect}, \texttt{top35}, \texttt{top45}, \texttt{top55}, \texttt{top65}, \texttt{top75}, and \texttt{top85}, and reports the empirical lower-tail probabilities of the induced upper-tail \(p\)-values on the select-stage empirical \(F_0\) subset.

The broad pool \texttt{all\_incorrect} is anti-conservative throughout. At the main setting \(N_{\mathrm{test}}=30\), it yields \(\widehat{\Pr}(p\le 0.10)=0.1136\) and \(\widehat{\Pr}(p\le 0.20)=0.4205\), well above the corresponding values for the harder upper-tail families. As the retained bank is restricted to stronger negatives, the induced upper tail becomes more conservative. This shift is useful only up to a point. More aggressive families suppress false release more strongly, but they also raise the baseline \(p\)-values and make feasible-side release harder. In the main \(N_{\mathrm{test}}=30\) regime, the families from \texttt{top35} through \texttt{top55} all yield \(\widehat{\Pr}(p\le 0.20)=0.1136\), while \texttt{top65} already increases this value to \(0.2159\). This is the numerical pattern behind the main-text choice of \texttt{top55}. It is the most relaxed family in this reduced grid that remains within the acceptable range under the held-out select-stage diagnosis.

% =========================================================
% Table S1: broader diagnosis (stacked vertical panels)
% =========================================================
\begin{table}[!h]
\centering
\caption{Broader select-stage diagnosis over candidate upper-tail pool families. For \(N_{\mathrm{test}}=10,20,30\), the select-stage empirical \(F_0\) subset contains \(10\) tasks and \(91\), \(90\), and \(88\) executable steps, respectively.}
\label{tab:supp_diag_compact}
\small

\textit{Panel A. \(N_{\mathrm{test}}=10\)}
\vspace{0.2em}

\begin{tabular}{lccccc}
\toprule
Family & Bank size & \(\widehat{\Pr}(p\le 0.05)\) & \(\widehat{\Pr}(p\le 0.10)\) & \(\widehat{\Pr}(p\le 0.20)\) & mean\((p)\) \\
\midrule
\texttt{all\_incorrect} & 317 & 0.0000 & 0.0000 & 0.1099 & 0.7099 \\
\texttt{top35}          & 115 & 0.0000 & 0.0000 & 0.0000 & 0.8511 \\
\texttt{top45}          & 147 & 0.0000 & 0.0000 & 0.0000 & 0.8144 \\
\texttt{top55}          & 179 & 0.0000 & 0.0000 & 0.0000 & 0.7907 \\
\texttt{top65}          & 221 & 0.0000 & 0.0000 & 0.1099 & 0.7602 \\
\texttt{top75}          & 317 & 0.0000 & 0.0000 & 0.1099 & 0.7099 \\
\texttt{top85}          & 317 & 0.0000 & 0.0000 & 0.1099 & 0.7099 \\
\bottomrule
\end{tabular}

\vspace{0.9em}
\textit{Panel B. \(N_{\mathrm{test}}=20\)}
\vspace{0.2em}

\begin{tabular}{lccccc}
\toprule
Family & Bank size & \(\widehat{\Pr}(p\le 0.05)\) & \(\widehat{\Pr}(p\le 0.10)\) & \(\widehat{\Pr}(p\le 0.20)\) & mean\((p)\) \\
\midrule
\texttt{all\_incorrect} & 304 & 0.0000 & 0.0000 & 0.1778 & 0.6839 \\
\texttt{top35}          & 109 & 0.0000 & 0.0000 & 0.0000 & 0.8545 \\
\texttt{top45}          & 147 & 0.0000 & 0.0000 & 0.0000 & 0.8149 \\
\texttt{top55}          & 169 & 0.0000 & 0.0000 & 0.1111 & 0.7971 \\
\texttt{top65}          & 197 & 0.0000 & 0.0000 & 0.1778 & 0.7746 \\
\texttt{top75}          & 231 & 0.0000 & 0.0000 & 0.1778 & 0.7523 \\
\texttt{top85}          & 304 & 0.0000 & 0.0000 & 0.1778 & 0.6839 \\
\bottomrule
\end{tabular}

\vspace{0.9em}
\textit{Panel C. \(N_{\mathrm{test}}=30\)}
\vspace{0.2em}

\begin{tabular}{lccccc}
\toprule
Family & Bank size & \(\widehat{\Pr}(p\le 0.05)\) & \(\widehat{\Pr}(p\le 0.10)\) & \(\widehat{\Pr}(p\le 0.20)\) & mean\((p)\) \\
\midrule
\texttt{all\_incorrect} & 309 & 0.0000 & 0.1136 & 0.4205 & 0.5649 \\
\texttt{top35}          & 127 & 0.0000 & 0.0000 & 0.1136 & 0.7110 \\
\texttt{top45}          & 150 & 0.0000 & 0.0000 & 0.1136 & 0.6893 \\
\texttt{top55}          & 170 & 0.0000 & 0.0000 & 0.1136 & 0.6744 \\
\texttt{top65}          & 207 & 0.0000 & 0.0000 & 0.2159 & 0.6444 \\
\texttt{top75}          & 238 & 0.0000 & 0.0000 & 0.3068 & 0.6247 \\
\texttt{top85}          & 309 & 0.0000 & 0.1136 & 0.4205 & 0.5649 \\
\bottomrule
\end{tabular}
\end{table}

\subsection{Robustness to practical design choices}
\label{subsec:mbpp_robustness}
%{\color{blue}
We examine the robustness of the proposed wrapper to two practical design choices: the resolution of the visible verifier, $N_{\mathrm{test}}$, and the betting exponent, $\eta$. Table~\ref{tab:robustness_compact} demonstrates that greater conservativeness in either dimension improves protection against false release, but it also delays or reduces feasible-side activity.

\begin{table}[h!]
\centering
\caption{Robustness to practical design choices for the proposed method under the selected \texttt{top55} family. Panel A varies verifier resolution at the main betting exponent \(\eta=0.7\). Panel B varies the betting exponent at the main verifier setting \(N_{\mathrm{test}}=30\). All entries are rounded to two decimal places.}
\label{tab:robustness_compact}
\small

\textit{Panel A. Verifier resolution at \(\eta=0.7\)}
\vspace{0.2em}

\begin{tabular}{ccccccc}
\toprule
\(N_{\mathrm{test}}\) & \(\alpha\) & \multicolumn{2}{c}{$F_0$} & \multicolumn{3}{c}{$F_1$} \\
\cmidrule{3-4} \cmidrule{5-7}
& & False-release & Release step & Release rate & Failure$\,|$release & Release step \\
\midrule
\multirow{3}{*}{10}
& 0.20 & 0.09 & 10.00 & 0.63 & 0.02 & 10.00 \\
& 0.10 & 0.00 & ---   & 0.00 & ---  & ---   \\
& 0.05 & 0.00 & ---   & 0.00 & ---  & ---   \\
\midrule
\multirow{3}{*}{20}
& 0.20 & 0.03 & 7.00  & 0.70 & 0.03 & 7.07  \\
& 0.10 & 0.00 & ---   & 0.65 & 0.01 & 9.04  \\
& 0.05 & 0.00 & ---   & 0.00 & ---  & ---   \\
\midrule
\multirow{3}{*}{30}
& 0.20 & 0.03 & 10.00 & 0.80 & 0.00 & 4.30  \\
& 0.10 & 0.00 & ---   & 0.77 & 0.00 & 6.27  \\
& 0.05 & 0.00 & ---   & 0.74 & 0.01 & 7.22  \\
\bottomrule
\end{tabular}

\vspace{0.9em}
\textit{Panel B. Betting exponent sensitivity at \(N_{\mathrm{test}}=30\)}
\vspace{0.2em}

\begin{tabular}{ccccccc}
\toprule
\(\eta\) & \(\alpha\) & \multicolumn{2}{c}{$F_0$} & \multicolumn{3}{c}{$F_1$} \\
\cmidrule{3-4} \cmidrule{5-7}
& & False-release & Release step & Release rate & Failure$\,|$release & Release step \\
\midrule
\multirow{3}{*}{0.3}
& 0.20 & 0.00 & ---   & 0.74 & 0.01 & 8.18  \\
& 0.10 & 0.00 & ---   & 0.00 & ---  & ---   \\
& 0.05 & 0.00 & ---   & 0.00 & ---  & ---   \\
\midrule
\multirow{3}{*}{0.5}
& 0.20 & 0.00 & ---   & 0.77 & 0.01 & 6.25  \\
& 0.10 & 0.00 & ---   & 0.68 & 0.01 & 8.02  \\
& 0.05 & 0.00 & ---   & 0.64 & 0.00 & 10.00 \\
\midrule
\multirow{3}{*}{0.7}
& 0.20 & 0.03 & 10.00 & 0.80 & 0.00 & 4.30  \\
& 0.10 & 0.00 & ---   & 0.77 & 0.00 & 6.27  \\
& 0.05 & 0.00 & ---   & 0.74 & 0.01 & 7.22  \\
\midrule
\multirow{3}{*}{0.9}
& 0.20 & 0.03 & 6.00  & 0.82 & 0.01 & 3.30  \\
& 0.10 & 0.03 & 9.00  & 0.82 & 0.02 & 4.37  \\
& 0.05 & 0.00 & ---   & 0.78 & 0.02 & 5.35  \\
\bottomrule
\end{tabular}
\end{table}

We first vary the verifier resolution with $N_{\mathrm{test}} \in \{10, 20, 30\}$, while keeping the selected reference-pool family fixed at \texttt{top55}. As Panel A shows, reducing $N_{\mathrm{test}}$ coarsens the discrete $p$-value scale and correspondingly weakens the strength of the calibrated evidence. At $N_{\mathrm{test}}=20$, the wrapper remains active at $\alpha=0.20$ and $0.10$, although feasible-side releases occur noticeably later than in the main $N_{\mathrm{test}}=30$ setting. At $N_{\mathrm{test}}=10$, the degradation is more pronounced: the resolution becomes so coarse that the accumulated wealth no longer reaches the stricter release thresholds ($\alpha \in \{0.10, 0.05\}$) within the ten-step horizon, rendering the method inactive on feasible tasks. The infeasible-side results move much less dramatically, indicating that a coarser verifier primarily reduces feasible-side utility rather than materially weakening false-release control.

We next vary the betting exponent $\eta \in \{0.3, 0.5, 0.7, 0.9\}$ under the selected \texttt{top55} family at the main verifier setting $N_{\mathrm{test}}=30$. Panel B shows a complementary trade-off. Smaller values produce a more conservative wealth process and can make the method inactive at stringent thresholds. Larger values lead to earlier and more frequent releases, but they also weaken the empirical safety margin on infeasible tasks. In particular, the aggressive choice $\eta=0.9$ accelerates release but incurs nonzero false release on $F_0$ already at $\alpha=0.10$, whereas $\eta=0.3$ is so conservative that it becomes inactive at $\alpha \in \{0.10, 0.05\}$. Among the values considered, $\eta=0.7$ provides the most stable compromise, maintaining substantial feasible-side activity while preserving strong infeasible-side control in the main setting. 

These robustness patterns are also consistent with Theorem~\ref{thm:feasible_power_cert_gain}. Reducing \(N_{\mathrm{test}}\) coarsens the verifier score and weakens the calibrated separation between correct candidates and hard negatives, making it harder for the cumulative gain to exceed \(A_{\alpha,T}\) within ten steps. Increasing \(\eta\) enlarges the gain scale \(Z_t\), which can improve feasible-side activity, but the same aggressiveness can reduce the empirical safety margin on \(F_0\).

These robustness results support the design choices used in the main analysis. A moderately resolved verifier is important for maintaining informative evidence accumulation, and \(\eta=0.7\) provides a stable empirical compromise between infeasible-side safety and feasible-side release.
%}

\subsection{Additional task-level trajectory}
\label{subsec:additional_traces}

%{\color{blue}
To complement the main trace analysis, Table~\ref{tab:supp_additional_trace} records one additional feasible-side trajectory whose verifier signal is high but not a monotone certificate of correctness. The example is useful precisely because it exposes the cost and the benefit of the conservative anytime-valid wrapper: it may not release at the first correct candidate, but when it does release, the selected candidate is correct.
%}

\begin{table}[!t]
\centering
\caption{{\color{black}\textbf{Additional task-level trace illustrating a conservative correct release under an oscillating feasible trajectory.} For \texttt{Mbpp/643}, high verifier scores occur for both correct and incorrect candidates. \texttt{Entropy} releases correctly at the first candidate, \texttt{Stability} releases incorrectly at the next high-score candidate, and the proposed wrapper releases later, at a correct candidate, after accumulated evidence crosses the anytime threshold.}}
\label{tab:supp_additional_trace}
\small

%{\color{blue}
\textit{Oscillating feasible-side success (\texttt{Mbpp/643})}
\vspace{0.2em}

\begin{tabular}{cccccl}
\toprule
Step & Score & Correct & \(p_t\) & Wealth \(E_t\) & Decision \\
\midrule
1  & 1.000 & T & 0.146 & 1.559  & \texttt{Entropy}: release (Correct) \\
2  & 0.967 & F & 0.216 & 1.848  & \texttt{Stability}: release (Error) \\
3  & 0.967 & F & 0.216 & 2.191  & --- \\
4  & 0.967 & F & 0.216 & 2.596  & --- \\
5  & 1.000 & T & 0.146 & 4.049  & --- \\
6  & 0.967 & F & 0.216 & 4.799  & --- \\
7  & 1.000 & T & 0.146 & 7.483  & --- \\
8  & 0.967 & F & 0.216 & 8.869  & --- \\
9  & 1.000 & T & 0.146 & 13.831 & \texttt{Ours}: release (Correct) \\
10 & 1.000 & T & 0.146 & 21.570 & \texttt{First\_p}: no release \\
\bottomrule
\end{tabular}
%}
\end{table}

%{\color{blue}
Table~\ref{tab:supp_additional_trace} studies \texttt{Mbpp/643}, which asks whether a string contains the letter `z' away from both word boundaries. Boundary-sensitive text-matching tasks admit many partially correct rules. A candidate may capture the broad pattern of interior occurrences while still mishandling edge cases involving initial or terminal positions, repeated \texttt{z}'s, or boundary-adjacent matches. This makes high visible scores possible even when the candidate logic is still incorrect.

This trajectory also illustrates the cost of our conservative decision rule. The first candidate is already oracle-correct, and \texttt{Entropy} releases correctly at this point. Our wrapper does not release immediately, because even a perfect visible score corresponds to a calibrated value \(p_t=0.146\) under the selected \texttt{top55} null pool, which is above the nominal \(\alpha=0.10\) one-step release threshold. This is an intentional consequence of calibrating against strong hard-negative candidates rather than treating a raw score of \(1.000\) as automatically decisive.

The same trajectory shows why this conservatism can be useful. Immediately after the first correct candidate, the next three candidates have high visible scores \(0.967\) but are incorrect under the hidden oracle. The \texttt{Stability} rule releases at \(t=2\) after a short high-score window, and that release is an error. Later, correctness continues to alternate: steps \(5\) and \(7\) are correct, step \(8\) is incorrect, and steps \(9\) and \(10\) are correct. Thus, local score smoothness and high raw verifier scores are informative but not decisive in this example.

The proposed wrapper responds differently. It does not claim the earliest correct candidate, but it continues to accumulate calibrated evidence across the trajectory. Since every single-step value remains above \(\alpha=0.10\), \texttt{First\_p} never releases. In contrast, the wealth process eventually crosses the anytime threshold \(1/\alpha=10\), reaching \(E_t=13.831\) at \(t=9\). The candidate selected at this first crossing time is oracle-correct. This example therefore highlights a central tradeoff of the method: conservative calibration can delay release even when an early candidate is correct, but the actual release event is tied to accumulated evidence and occurs here on a true candidate rather than on the intervening high-score failures.
%}

{\color{black}
\subsection{Certifiable-gain diagnostics for feasible-side release}
\label{subsec:sim_theory}

Theorem~\ref{thm:feasible_power_cert_gain} identifies calibrated gain from correct candidates as the central feasible-side quantity. The cached MBPP+ trajectories allow this quantity to be measured post hoc, because each candidate has both a calibrated verifier-derived \(p\)-value and a hidden-oracle correctness label. We therefore compute trajectory-level gain decompositions that separate the evidence observed by the wrapper from the portion attributable to oracle-correct candidates.

\subsubsection{Trajectory-level gain decomposition}

For a fixed betting function \(f\), define the calibrated gain \(Z_t=\log f(p_t)-\log f(1)\), the cumulative observed gain \(G_t=\sum_{j=1}^t Z_j\), the cumulative correct-candidate gain \(G_t^+=\sum_{j=1}^t Y_jZ_j\), and the required cumulative gain \(A_{\alpha,t}=\log(1/\alpha)-t\log f(1)\). The release condition satisfies
\[
E_t\ge 1/\alpha
\quad\Longleftrightarrow\quad
G_t\ge A_{\alpha,t}.
\]
The wrapper observes \(G_t\), since it is determined by the online calibrated \(p\)-values. The quantity \(G_t^+\) is a post hoc attribution using hidden-oracle labels, measuring the portion of observed gain contributed by oracle-correct candidates. It is not used by the wrapper during deployment.

We also record the maximum observed margin and maximum correct-candidate margin,
\[
M(x):=\max_{t\le T_{\max}}\{G_t(x)-A_{\alpha,t}\},
\qquad
M^+(x)
:=
\max_{t\le T_{\max}}\{G_t^+(x)-A_{\alpha,t}\}.
\]
These margins summarize whether a trajectory ever supplies enough observed gain, or enough correct-candidate gain, to exceed the required gain curve. In the notation of Theorem~\ref{thm:feasible_power_cert_gain}, \(B_T\) is a conditional expected gain lower bound rather than an observed trajectory-level quantity. The diagnostics \(G_t^+\) and \(M^+\) provide empirical, post hoc analogues for examining the mechanism behind feasible-side release. All quantities in this subsection are computed from the cached trajectories without regenerating model outputs or rerunning tests.

\begin{figure}[!htbp]
\centering
\includegraphics[width=0.98\textwidth]{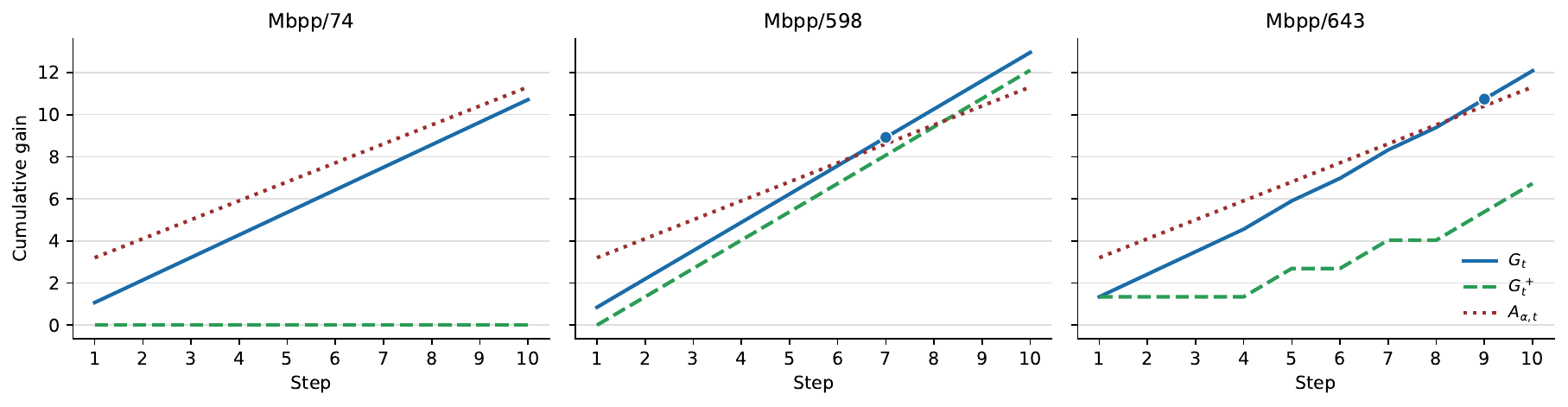}
\caption{\textbf{Representative gain trajectories.} The solid curve is the observed cumulative calibrated gain \(G_t\), the dashed curve is the hidden-oracle correct-candidate gain \(G_t^+\), and the dotted curve is the required gain \(A_{\alpha,t}\) for the main setting \(\alpha=0.10,\eta=0.7,N_{\mathrm{test}}=30\). Markers denote the first release time of the proposed wrapper.}
\label{fig:representative_gain_trajectories}
\end{figure}

\begin{table}[!htbp]
\centering
\caption{\textbf{Representative trajectory gain summaries.} All entries use \(\alpha=0.10\), \(\eta=0.7\), \(N_{\mathrm{test}}=30\), and the selected \texttt{top55} hard-negative pool. Here \(M=\max_t(G_t-A_{\alpha,t})\) and \(M^+=\max_t(G_t^+-A_{\alpha,t})\).}
\label{tab:representative_gain_summary}
\small
\begin{tabular}{llrrrrr}
\toprule
Task & Release & \(G_T\) & \(G_T^+\) & \(A_{\alpha,T}\) & \(M\) & \(M^+\) \\
\midrule
\texttt{Mbpp/74}  & ---         & 10.72 & 0.00  & 11.32 & -0.60 & -3.20 \\
\texttt{Mbpp/598} & 7 (correct) & 12.96 & 12.11 & 11.32 &  1.64 &  0.79 \\
\texttt{Mbpp/643} & 9 (correct) & 12.09 & 6.73  & 11.32 &  0.77 & -1.86 \\
\bottomrule
\end{tabular}
\end{table}

Figure~\ref{fig:representative_gain_trajectories} and Table~\ref{tab:representative_gain_summary} show three distinct gain patterns. For \texttt{Mbpp/74}, no candidate is correct under the hidden oracle, so \(G_t^+=0\) throughout. The raw verifier scores are persistently high enough to generate some observed gain, but \(G_T=10.72\) remains below the required terminal gain \(A_{\alpha,T}=11.32\), and the maximum observed margin is negative. This is the gain-level representation of safe abstention.

For \texttt{Mbpp/598}, the hidden oracle labels the candidates correct from the second step onward. The correct-candidate gain therefore rises nearly in parallel with the observed gain, with \(G_T^+=12.11\) accounting for \(93.5\%\) of terminal observed gain. The observed path crosses the threshold at \(t=7\), and the selected candidate is correct. This is the cleanest instance of the mechanism in Theorem~\ref{thm:feasible_power_cert_gain}: individually moderate calibrated evidence becomes decisive after repeated correct candidates.

The \texttt{Mbpp/643} trajectory is more nuanced and complements the trace in Table~\ref{tab:supp_additional_trace}. The wrapper releases correctly at \(t=9\), but \(G_T^+=6.73\) is well below \(A_{\alpha,T}=11.32\), and \(M^+<0\). Thus the crossing is not explained by correct-candidate gain alone; high-scoring incorrect candidates also contribute observed calibrated evidence. This distinction is substantively informative: \(G_t\) is the online evidence process used by the wrapper, while \(G_t^+\) is the hidden-oracle attribution of that evidence. The release event is still tied to a valid threshold crossing, and in this trajectory the first crossing occurs at a true candidate, but the path exposes the cost and structure of conservative calibrated stopping in an oscillating feasible task.

\FloatBarrier

\subsubsection{Held-out feasible-side gain diagnostics}

The held-out feasible-side summaries in Figure~\ref{fig:f1_margin_diagnostics} and Table~\ref{tab:f1_margin_by_release} show that release activity is strongly aligned with the accumulation of correct-candidate gain.

\begin{figure}[!htbp]
\centering
\includegraphics[width=0.86\textwidth]{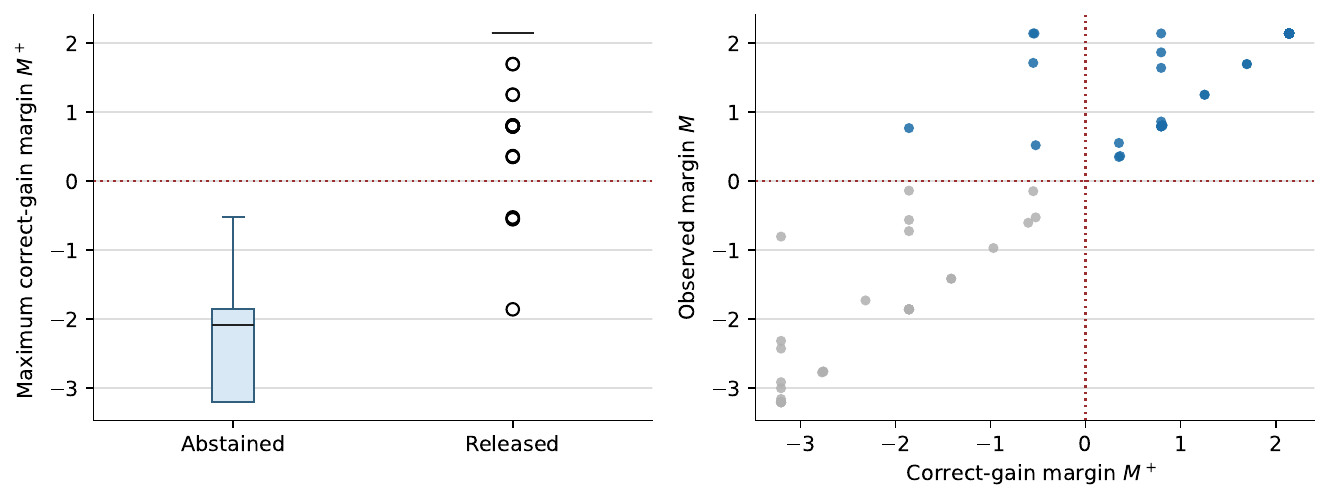}
\caption{\textbf{Feasible-side margin diagnostics.} Left: distribution of the maximum correct-gain margin \(M^+\) among feasible held-out tasks, split by whether the wrapper releases in the main setting. Right: observed margin \(M\) versus correct-gain margin \(M^+\). Blue points are released tasks and gray points are abstentions.}
\label{fig:f1_margin_diagnostics}
\end{figure}

\begin{table}[!htbp]
\centering
\caption{\textbf{Held-out feasible-side margin summaries.} The main setting contains 148 empirical feasible tasks. Released tasks are those for which the wrapper crosses the anytime threshold within ten steps.}
\label{tab:f1_margin_by_release}
\small
\begin{tabular}{lrrrrrr}
\toprule
Group & \(n\) & mean \(M\) & median \(M\) & mean \(M^+\) & median \(M^+\) & \(\Pr(M^+>0)\) \\
\midrule
Released  & 114 &  1.93 &  2.14 &  1.80 &  2.14 & 0.96 \\
Abstained &  34 & -2.01 & -1.86 & -2.30 & -2.09 & 0.00 \\
\bottomrule
\end{tabular}
\end{table}

Among the 148 feasible tasks in the main setting, the wrapper releases on 114. These released tasks have mean observed margin \(1.93\) and mean correct-gain margin \(1.80\), whereas the 34 abstained feasible tasks have mean margins \(-2.01\) and \(-2.30\), respectively. Moreover, \(95.6\%\) of released feasible tasks have \(M^+>0\), compared with \(0\%\) of abstained feasible tasks. The small difference between \(M\) and \(M^+\) for the released group also indicates that, in aggregate, the observed threshold crossings are mostly supplied by candidates that are correct under the hidden oracle.

\begin{table}[!htbp]
\centering
\caption{\textbf{Stepwise decomposition of feasible-side gain.} Estimates are computed over the 148 empirical feasible tasks in the main setting. The column \(\widehat{\pi}_t\) is the fraction of tasks with a correct candidate at step \(t\), \(\widehat{\bar Z}_t\) is the mean calibrated gain among correct candidates, and \(\widehat{\mathbb E}(Y_tZ_t)\) is their product as observed in the held-out trajectories.}
\label{tab:step_gain_decomposition}
\small
\begin{tabular}{rrrrrr}
\toprule
\(t\) & \(\widehat{\pi}_t\) & \(\widehat{\bar Z}_t\) & \(\widehat{\mathbb E}(Y_tZ_t)\) & \(\sum_{j\le t}\widehat{\mathbb E}(Y_jZ_j)\) & \(A_{\alpha,t}\) \\
\midrule
1  & 0.85 & 1.34 & 1.14 &  1.14 &  3.20 \\
2  & 0.82 & 1.34 & 1.10 &  2.24 &  4.11 \\
3  & 0.84 & 1.34 & 1.13 &  3.37 &  5.01 \\
4  & 0.83 & 1.34 & 1.12 &  4.49 &  5.91 \\
5  & 0.82 & 1.34 & 1.10 &  5.58 &  6.81 \\
6  & 0.86 & 1.34 & 1.16 &  6.75 &  7.71 \\
7  & 0.86 & 1.34 & 1.15 &  7.90 &  8.61 \\
8  & 0.82 & 1.34 & 1.11 &  9.01 &  9.52 \\
9  & 0.84 & 1.34 & 1.13 & 10.14 & 10.42 \\
10 & 0.81 & 1.34 & 1.09 & 11.23 & 11.32 \\
\bottomrule
\end{tabular}
\end{table}

Table~\ref{tab:step_gain_decomposition} makes the decomposition in Theorem~\ref{thm:feasible_power_cert_gain} concrete. The empirical correct-candidate frequency is high throughout the ten-step horizon, ranging from \(0.81\) to \(0.86\). Conditional on correctness, the mean calibrated gain is stable at about \(1.34\), reflecting the fact that correct candidates are typically assigned near-perfect verifier scores under the \texttt{top55} calibration. The resulting cumulative mean correct gain reaches \(11.23\) by \(T=10\), nearly matching the required terminal gain \(A_{\alpha,T}=11.32\). This near equality explains why the main setting is active but not overly aggressive: feasible tasks often supply enough calibrated gain to cross the threshold, but there remains a nontrivial abstention region when the correct-candidate gain arrives too late or too weakly.

\FloatBarrier

\subsubsection{Gain sensitivity to verifier resolution and betting strength}
The sensitivity results in Figure~\ref{fig:gain_sensitivity} and Table~\ref{tab:gain_sensitivity} align with the two sources of feasible-side power highlighted by Theorem~\ref{thm:feasible_power_cert_gain}. 

\begin{figure}[!htbp]
\centering
\includegraphics[width=0.88\textwidth]{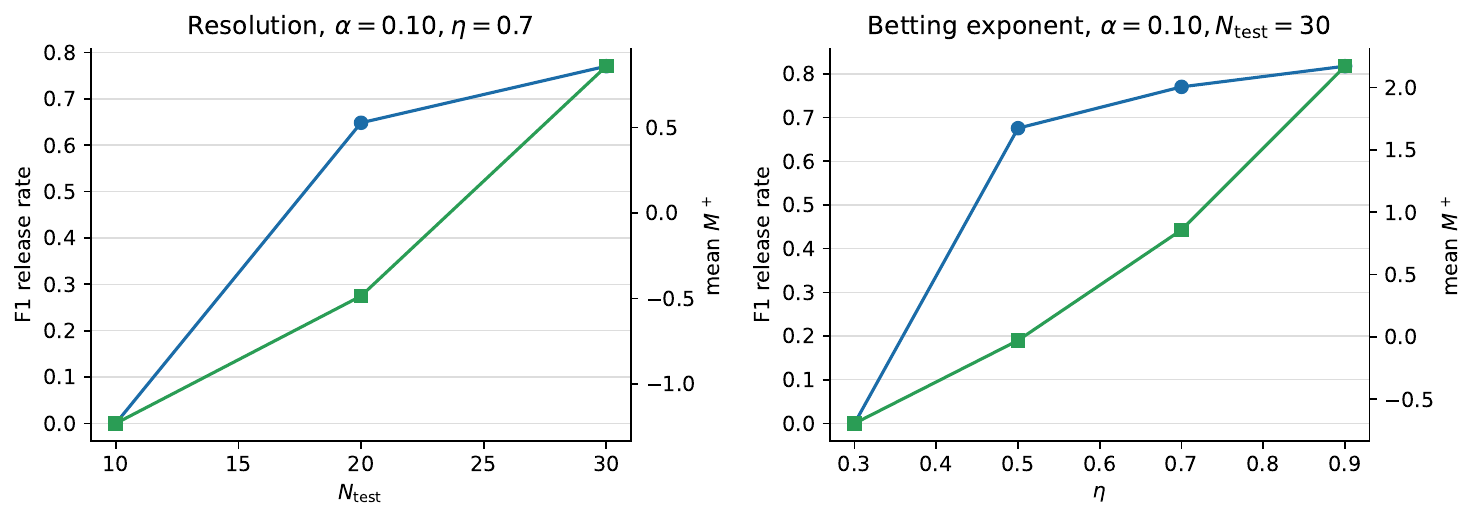}
\caption{\textbf{Gain sensitivity.} Left: feasible-side release rate and mean correct-gain margin as the verifier resolution changes, with \(\alpha=0.10\) and \(\eta=0.7\). Right: the same quantities as the betting exponent changes, with \(\alpha=0.10\) and \(N_{\mathrm{test}}=30\).}
\label{fig:gain_sensitivity}
\end{figure}

\begin{table}[!htbp]
\centering
\caption{\textbf{Feasible-side gain sensitivity at \(\alpha=0.10\).} Panel A varies verifier resolution with \(\eta=0.7\). Panel B varies betting exponent with \(N_{\mathrm{test}}=30\).}
\label{tab:gain_sensitivity}
\small
\begin{tabular}{lrrrrr}
\toprule
Setting & Release rate & Failure\(\mid\)release & Release step & mean \(G_T^+\) & mean \(M^+\) \\
\midrule
\multicolumn{6}{l}{\textit{Panel A. Varying \(N_{\mathrm{test}}\), \(\eta=0.7\)}}\\
\(N_{\mathrm{test}}=10\) & 0.00 & ---  & ---  &  9.00 & -1.23 \\
\(N_{\mathrm{test}}=20\) & 0.65 & 0.01 & 9.04 &  9.91 & -0.49 \\
\(N_{\mathrm{test}}=30\) & 0.77 & 0.00 & 6.27 & 11.23 &  0.86 \\
\midrule
\multicolumn{6}{l}{\textit{Panel B. Varying \(\eta\), \(N_{\mathrm{test}}=30\)}}\\
\(\eta=0.3\) & 0.00 & ---  & ---  &  4.81 & -0.69 \\
\(\eta=0.5\) & 0.68 & 0.01 & 8.02 &  8.02 & -0.03 \\
\(\eta=0.7\) & 0.77 & 0.00 & 6.27 & 11.23 &  0.86 \\
\(\eta=0.9\) & 0.82 & 0.02 & 4.37 & 14.44 &  2.17 \\
\bottomrule
\end{tabular}
\end{table}

Increasing \(N_{\mathrm{test}}\) sharpens the calibrated separation between correct candidates and the hard-negative pool. At \(\alpha=0.10\) and \(\eta=0.7\), the mean terminal correct gain increases from \(9.00\) at \(N_{\mathrm{test}}=10\) to \(11.23\) at \(N_{\mathrm{test}}=30\), and the mean correct-gain margin moves from \(-1.23\) to \(0.86\). The feasible-side release rate correspondingly rises from \(0\) to \(0.77\). Varying \(\eta\) changes the gain scale directly. At \(N_{\mathrm{test}}=30\), increasing \(\eta\) from \(0.3\) to \(0.9\) raises mean \(G_T^+\) from \(4.81\) to \(14.44\) and accelerates release. This power gain is not free: as shown in the main robustness table, the most aggressive setting \(\eta=0.9\) also introduces nonzero false release on \(F_0\) at \(\alpha=0.10\). The selected \(\eta=0.7\) therefore sits in the empirical region where correct-candidate gain is large enough for substantial feasible-side activity while preserving the stronger infeasible-side safety margin observed in the main setting.
}

\section{Proofs}\label{app:proofs}

This section presents detailed proofs for the theoretical results established in the main text, organized in the order of their appearance.

{\color{black}
\subsection{Proof of Proposition~\ref{prop:rout_from_primitive}}

If \(\mathbb P(D=1)=0\), the release-conditional quantity is vacuous. We therefore consider the nondegenerate case \(\mathbb P(D=1)>0\). Write
\[
q:=\mathbb P(Y_\tau=0\mid D=1,F=1),
\qquad
\beta:=\mathbb P(D=1\mid F=1).
\]
If \(\mathbb P(D=1,F=1)=0\), \(q\) may be defined arbitrarily in \([0,1]\), since it is multiplied below by \(\mathbb P(F=1\mid D=1)=0\).

\noindent\textbf{Step 1. Decompose the release-conditional failure rate by feasibility.}
By the law of total probability conditional on \(\{D=1\}\),
\[
\mathbb P(Y_\tau=0\mid D=1)
=
\mathbb P(Y_\tau=0\mid D=1,F=0)\mathbb P(F=0\mid D=1)
+
q\,\mathbb P(F=1\mid D=1).
\]
Also,
\[
\left\{
\begin{aligned}
\mathbb P(Y_\tau=0\mid D=1,F=0)&\le 1,\\
\mathbb P(F=1\mid D=1)&=1-\mathbb P(F=0\mid D=1).
\end{aligned}
\right.
\]
Substituting these two facts into the preceding display gives
\[
\mathbb P(Y_\tau=0\mid D=1)
\le
\mathbb P(F=0\mid D=1)
+
q\{1-\mathbb P(F=0\mid D=1)\}.
\]
Rearranging the right-hand side,
\[
\mathbb P(F=0\mid D=1)
+
q\{1-\mathbb P(F=0\mid D=1)\}
=
q+(1-q)\mathbb P(F=0\mid D=1).
\]
Thus
\[
R_{\mathrm{out}}
=
\mathbb P(Y_\tau=0\mid D=1)
\le
q+(1-q)\mathbb P(F=0\mid D=1).
\]

\noindent\textbf{Step 2. Bound the posterior probability of infeasibility after release.}
By Bayes' rule,
\[
\mathbb P(F=0\mid D=1)
=
\frac{\mathbb P(D=1\mid F=0)\mathbb P(F=0)}
{\mathbb P(D=1)}.
\]
Using \(\pi_0=\mathbb P(F=0)\) and decomposing \(\mathbb P(D=1)\) over \(F\in\{0,1\}\),
\[
\mathbb P(D=1)
=
\mathbb P(D=1\mid F=0)\mathbb P(F=0)
+
\mathbb P(D=1\mid F=1)\mathbb P(F=1).
\]
Since \(\mathbb P(F=1)=1-\pi_0\) and \(\beta=\mathbb P(D=1\mid F=1)\), this becomes
\[
\mathbb P(D=1)
=
\pi_0\mathbb P(D=1\mid F=0)
+
(1-\pi_0)\beta.
\]
Substituting this denominator into Bayes' rule gives
\[
\mathbb P(F=0\mid D=1)
=
\frac{\pi_0\mathbb P(D=1\mid F=0)}
{\pi_0\mathbb P(D=1\mid F=0)+(1-\pi_0)\beta}.
\]
Let \(x_0:=\mathbb P(D=1\mid F=0)\). The assumed false-release guarantee gives \(x_0\le\alpha\).

If \(\beta=0\), then \(\mathbb P(D=1)>0\) implies \(\pi_0x_0>0\), so \(\mathbb P(F=0\mid D=1)=1\). In this case,
\[
\frac{\pi_0\alpha}{\pi_0\alpha+(1-\pi_0)\beta}
=
\frac{\pi_0\alpha}{\pi_0\alpha}
=
1,
\]
and the desired bound on \(\mathbb P(F=0\mid D=1)\) holds. We now consider \(\beta>0\).

Define
\[
g(x):=\frac{\pi_0x}{\pi_0x+(1-\pi_0)\beta},
\qquad x\ge0.
\]
The denominator is positive for all \(x\ge0\). Differentiating,
\[
g'(x)
=
\frac{
\pi_0\{\pi_0x+(1-\pi_0)\beta\}
-
\pi_0x\pi_0
}{
\{\pi_0x+(1-\pi_0)\beta\}^2
}.
\]
Expanding the numerator,
\[
\pi_0\{\pi_0x+(1-\pi_0)\beta\}
-
\pi_0x\pi_0
=
\pi_0^2x+\pi_0(1-\pi_0)\beta-\pi_0^2x
=
\pi_0(1-\pi_0)\beta.
\]
Hence
\[
g'(x)
=
\frac{\pi_0(1-\pi_0)\beta}
{\{\pi_0x+(1-\pi_0)\beta\}^2}
\ge0.
\]
Thus \(g\) is non-decreasing. Since \(x_0\le\alpha\),
\[
\mathbb P(F=0\mid D=1)
=
g(x_0)
\le
g(\alpha)
=
\frac{\pi_0\alpha}{\pi_0\alpha+(1-\pi_0)\beta}.
\]

\noindent\textbf{Step 3. Substitute the infeasibility bound into the risk decomposition.}
From Step 1,
\[
R_{\mathrm{out}}
\le
q+(1-q)\mathbb P(F=0\mid D=1).
\]
Using the bound from Step 2,
\[
R_{\mathrm{out}}
\le
q
+
(1-q)
\frac{\pi_0\alpha}{\pi_0\alpha+(1-\pi_0)\beta}.
\]
Finally, substituting back
\[
\left\{
\begin{aligned}
q&=\mathbb P(Y_\tau=0\mid D=1,F=1),\\
\beta&=\mathbb P(D=1\mid F=1),
\end{aligned}
\right.
\]
we obtain
\[
\begin{split}
R_{\mathrm{out}}
&\le
\mathbb P(Y_\tau=0\mid D=1,F=1)\\
&\quad+
\Bigl(1-\mathbb P(Y_\tau=0\mid D=1,F=1)\Bigr)
\frac{\pi_0\alpha}
{\pi_0\alpha+(1-\pi_0)\mathbb P(D=1\mid F=1)}.
\end{split}
\]
This is exactly \eqref{eq:rout_translation}. \qed
}

{\color{black}
\subsection{Proof of Proposition~\ref{thm:optstop_ct}}
\label{proof:thm-optstop-ct-updated}

Fix \(T\in\mathbb N\). Throughout the proof, probabilities and expectations are taken under the conditional law given \(F=0\). For readability, write \(\mathbb P_0(\cdot):=\mathbb P(\cdot\mid F=0)\) and \(\mathbb E_0(\cdot):=\mathbb E(\cdot\mid F=0)\).

\noindent\textbf{Step 1. Define the survival events.}
Let \(A_0:=\Omega\) and, for \(t\ge1\), let \(A_t:=\{\tau_{\mathrm{naive}}>t\}\). Since
\[
\tau_{\mathrm{naive}}=\inf\{s\ge1:p_s\le\alpha\},
\]
the event \(\{\tau_{\mathrm{naive}}>t\}\) is exactly the event that no crossing has occurred through time \(t\). Hence
\[
A_t
=
\bigcap_{s=1}^t\{p_s>\alpha\}.
\]
In particular, \(A_t\in\mathcal G_t\), and \(A_{t-1}\in\mathcal G_{t-1}\) for every \(t\ge1\). Also,
\[
A_t=A_{t-1}\cap\{p_t>\alpha\}.
\]

\noindent\textbf{Step 2. Derive a one-step survival recursion.}
The assumption gives, for every \(t\ge1\),
\[
\mathbb P(p_t\le\alpha\mid\mathcal G_{t-1},F=0)\ge c_t
\qquad\text{almost surely}.
\]
Since \(\{p_t>\alpha\}\) is the complement of \(\{p_t\le\alpha\}\),
\[
\mathbb P(p_t>\alpha\mid\mathcal G_{t-1},F=0)
=
1-\mathbb P(p_t\le\alpha\mid\mathcal G_{t-1},F=0)
\le
1-c_t.
\]
Now use \(A_t=A_{t-1}\cap\{p_t>\alpha\}\). We have
\[
\mathbb P_0(A_t)
=
\mathbb E_0\!\left[
\mathbf 1\{A_t\}
\right]
=
\mathbb E_0\!\left[
\mathbf 1\{A_{t-1}\}\mathbf 1\{p_t>\alpha\}
\right].
\]
Taking conditional expectation given \(\mathcal G_{t-1}\),
\[
\mathbb E_0\!\left[
\mathbf 1\{A_{t-1}\}\mathbf 1\{p_t>\alpha\}
\right]
=
\mathbb E_0\!\left[
\mathbb E_0\!\left\{
\mathbf 1\{A_{t-1}\}\mathbf 1\{p_t>\alpha\}
\,\middle|\,
\mathcal G_{t-1}
\right\}
\right].
\]
Because \(A_{t-1}\in\mathcal G_{t-1}\), the indicator \(\mathbf 1\{A_{t-1}\}\) is \(\mathcal G_{t-1}\)-measurable, so
\[
\mathbb E_0\!\left\{
\mathbf 1\{A_{t-1}\}\mathbf 1\{p_t>\alpha\}
\,\middle|\,
\mathcal G_{t-1}
\right\}
=
\mathbf 1\{A_{t-1}\}
\mathbb E_0\!\left[
\mathbf 1\{p_t>\alpha\}
\,\middle|\,
\mathcal G_{t-1}
\right].
\]
The conditional expectation of the indicator is the conditional probability:
\[
\mathbb E_0\!\left[
\mathbf 1\{p_t>\alpha\}
\,\middle|\,
\mathcal G_{t-1}
\right]
=
\mathbb P(p_t>\alpha\mid\mathcal G_{t-1},F=0).
\]
Therefore,
\[
\mathbb P_0(A_t)
=
\mathbb E_0\!\left[
\mathbf 1\{A_{t-1}\}
\mathbb P(p_t>\alpha\mid\mathcal G_{t-1},F=0)
\right].
\]
Using the one-step bound \(\mathbb P(p_t>\alpha\mid\mathcal G_{t-1},F=0)\le1-c_t\),
\[
\mathbb P_0(A_t)
\le
\mathbb E_0\!\left[
\mathbf 1\{A_{t-1}\}(1-c_t)
\right].
\]
Since \(c_t\) is deterministic,
\[
\mathbb E_0\!\left[
\mathbf 1\{A_{t-1}\}(1-c_t)
\right]
=
(1-c_t)\mathbb E_0[\mathbf 1\{A_{t-1}\}]
=
(1-c_t)\mathbb P_0(A_{t-1}).
\]
Thus
\[
\mathbb P_0(A_t)\le(1-c_t)\mathbb P_0(A_{t-1}).
\]

\noindent\textbf{Step 3. Iterate the recursion.}
Applying the recursion at times \(T,T-1,\ldots,1\),
\[
\mathbb P_0(A_T)
\le
(1-c_T)\mathbb P_0(A_{T-1}).
\]
Applying the same inequality to \(\mathbb P_0(A_{T-1})\),
\[
\mathbb P_0(A_{T-1})
\le
(1-c_{T-1})\mathbb P_0(A_{T-2}).
\]
Substituting this into the previous display gives
\[
\mathbb P_0(A_T)
\le
(1-c_T)(1-c_{T-1})\mathbb P_0(A_{T-2}).
\]
Continuing in this way yields
\[
\mathbb P_0(A_T)
\le
\prod_{t=1}^T(1-c_t)\,\mathbb P_0(A_0).
\]
Since \(A_0=\Omega\), we have \(\mathbb P_0(A_0)=1\). Hence
\[
\mathbb P_0(A_T)
\le
\prod_{t=1}^T(1-c_t).
\]
Because \(A_T=\{\tau_{\mathrm{naive}}>T\}\),
\[
\mathbb P(\tau_{\mathrm{naive}}>T\mid F=0)
\le
\prod_{t=1}^T(1-c_t).
\]

\noindent\textbf{Step 4. Convert the product bound into the stated exponential bound.}
For every \(x\in[0,1]\), \(1-x\le e^{-x}\). Since \(c_t\in[0,1]\),
\[
1-c_t\le e^{-c_t}
\qquad\text{for each }t.
\]
Multiplying these inequalities over \(t=1,\ldots,T\),
\[
\prod_{t=1}^T(1-c_t)
\le
\prod_{t=1}^T e^{-c_t}.
\]
Using \(\prod_{t=1}^T e^{-c_t}=e^{-\sum_{t=1}^T c_t}\), we get
\[
\prod_{t=1}^T(1-c_t)
\le
\exp\!\left(-\sum_{t=1}^T c_t\right).
\]
Combining this with Step 3,
\[
\mathbb P(\tau_{\mathrm{naive}}>T\mid F=0)
\le
\exp\!\left(-\sum_{t=1}^T c_t\right).
\]
Taking complements,
\[
\mathbb P(\tau_{\mathrm{naive}}\le T\mid F=0)
=
1-\mathbb P(\tau_{\mathrm{naive}}>T\mid F=0).
\]
Therefore,
\[
\mathbb P(\tau_{\mathrm{naive}}\le T\mid F=0)
\ge
1-\exp\!\left(-\sum_{t=1}^T c_t\right).
\]

\noindent\textbf{Step 5. Infinite-horizon consequence.}
Assume now that \(\sum_{t\ge1}c_t=\infty\). From Step 4,
\[
\mathbb P(\tau_{\mathrm{naive}}>T\mid F=0)
\le
\exp\!\left(-\sum_{t=1}^T c_t\right).
\]
As \(T\to\infty\), the partial sums \(\sum_{t=1}^T c_t\) diverge to infinity, so
\[
\exp\!\left(-\sum_{t=1}^T c_t\right)\to0.
\]
The events \(\{\tau_{\mathrm{naive}}>T\}\) decrease to \(\{\tau_{\mathrm{naive}}=\infty\}\). By continuity from above,
\[
\mathbb P(\tau_{\mathrm{naive}}=\infty\mid F=0)
=
\lim_{T\to\infty}
\mathbb P(\tau_{\mathrm{naive}}>T\mid F=0).
\]
Using the bound above, the limit is zero:
\[
\mathbb P(\tau_{\mathrm{naive}}=\infty\mid F=0)=0.
\]
Thus
\[
\mathbb P(\tau_{\mathrm{naive}}<\infty\mid F=0)
=
1-\mathbb P(\tau_{\mathrm{naive}}=\infty\mid F=0)
=
1.
\]
This completes the proof. \qed
}

{\color{black}
\subsection{Proof of Theorem~\ref{thm:stepwise_pvalue}}

Fix \(t\ge1\). By the countability of \(\mathbb Q\), we may work on a full-probability event on which Assumption~\ref{assump:dom} holds simultaneously for every rational threshold \(q\in\mathbb Q\). Thus, on this event, for every \(q\in\mathbb Q\),
\[
\mathbb P\!\left(S_t\ge q\,\middle|\,\widetilde{\mathcal G}_{t-1},F=0\right)
\le
\frac{1}{n}\sum_{i=1}^n \mathbf 1\{R_i\ge q\}.
\]
Fix a realization of \(\widetilde{\mathcal G}_{t-1}\) on this event and work under the corresponding conditional law given \(F=0\). Under this conditional law, the reference scores \(R_1,\ldots,R_n\) are fixed numbers. Let \(N(s):=\sum_{i=1}^n\mathbf 1\{R_i\ge s\}\), so that \(p_t=(1+N(S_t))/(n+1)\). Let \(R_{(1)}\ge R_{(2)}\ge\cdots\ge R_{(n)}\) denote the reference scores sorted in non-increasing order, with ties kept with their multiplicities.

We prove that, for an arbitrary \(u\in[0,1]\),
\[
\mathbb P\!\left(p_t\le u\,\middle|\,\widetilde{\mathcal G}_{t-1},F=0\right)
\le u.
\]

\noindent\textbf{Step 1. Trivial endpoint cases.}
If \(u<1/(n+1)\), then \(p_t\ge1/(n+1)\) almost surely, and hence
\[
\mathbb P\!\left(p_t\le u\,\middle|\,\widetilde{\mathcal G}_{t-1},F=0\right)=0\le u.
\]
If \(u=1\), the claim is immediate. It remains to consider \(u\in[1/(n+1),1)\). Set \(k:=\lfloor (n+1)u-1\rfloor\). Then \(k\in\{0,1,\ldots,n-1\}\).

\noindent\textbf{Step 2. Reducing the small-\(p\) event to an order-statistic tail event.}
On the event \(\{p_t\le u\}\), we have
\[
\frac{1+N(S_t)}{n+1}\le u.
\]
Equivalently, \(1+N(S_t)\le (n+1)u\), so \(N(S_t)\le (n+1)u-1\). Since \(N(S_t)\) is integer-valued and \(k=\lfloor (n+1)u-1\rfloor\), this implies \(N(S_t)\le k\).

We now show that \(N(S_t)\le k\) forces \(S_t>R_{(k+1)}\). Suppose instead that \(S_t\le R_{(k+1)}\). Then the \(k+1\) reference scores \(R_{(1)},\ldots,R_{(k+1)}\) are all at least \(S_t\). Hence \(N(S_t)\ge k+1\), contradicting \(N(S_t)\le k\). Therefore,
\[
\{p_t\le u\}\subseteq \{S_t>R_{(k+1)}\}.
\]

\noindent\textbf{Step 3. Applying dominance through rational thresholds.}
Let \(a:=R_{(k+1)}\). Choose a sequence of rational numbers \(q_m\downarrow a\) with \(q_m>a\) for every \(m\). Since \(q_m>a=R_{(k+1)}\), at most \(k\) reference scores are at least \(q_m\). Thus \(N(q_m)\le k\). Applying Assumption~\ref{assump:dom} to the rational threshold \(q_m\) gives
\[
\mathbb P\!\left(S_t\ge q_m\,\middle|\,\widetilde{\mathcal G}_{t-1},F=0\right)
\le
\frac{N(q_m)}{n}
\le
\frac{k}{n}.
\]
As \(m\to\infty\), the events \(\{S_t\ge q_m\}\) increase to \(\{S_t>a\}\). By continuity from below,
\[
\mathbb P\!\left(S_t>a\,\middle|\,\widetilde{\mathcal G}_{t-1},F=0\right)
=
\lim_{m\to\infty}
\mathbb P\!\left(S_t\ge q_m\,\middle|\,\widetilde{\mathcal G}_{t-1},F=0\right)
\le
\frac{k}{n}.
\]

\noindent\textbf{Step 4. Completing the super-uniformity bound.}
From Step 2, \(\{p_t\le u\}\subseteq\{S_t>a\}\). Combining this inclusion with Step 3 gives
\[
\mathbb P\!\left(p_t\le u\,\middle|\,\widetilde{\mathcal G}_{t-1},F=0\right)
\le
\frac{k}{n}.
\]
Since \(k=\lfloor (n+1)u-1\rfloor\), we have \(k\le (n+1)u-1\). Therefore,
\[
\frac{k}{n}
\le
\frac{(n+1)u-1}{n}
=
u+\frac{u-1}{n}.
\]
Because \(u<1\) in the remaining case, \((u-1)/n\le0\), and hence
\[
\frac{k}{n}\le u.
\]
Thus
\[
\mathbb P\!\left(p_t\le u\,\middle|\,\widetilde{\mathcal G}_{t-1},F=0\right)
\le u.
\]
Together with the endpoint cases in Step 1, this proves the desired inequality for every \(u\in[0,1]\). Since the argument is pointwise on a full-probability event, the conditional super-uniformity statement holds almost surely. This proves \eqref{eq:su}. \qed
}

{\color{black}
\subsection{Proof of Proposition~\ref{prop:betting_martingale}}
\label{proof:prop_betting_martingale_updated}

Fix \(t\ge1\), and write \(\mathcal H_t:=\widetilde{\mathcal G}_t\). Throughout the proof, probabilities and expectations are taken under the conditional law given \(F=0\). We work on a full-probability event on which the conditional super-uniformity condition \eqref{eq:su} holds for all \(u\in[0,1]\). This is without loss, since it is enough to take the countable event on which \eqref{eq:su} holds for all rational \(u\), and then extend to all \(u\) by monotonicity and right-continuity of distribution functions.

\noindent\textbf{Step 1. Reduce the e-process condition to a one-step p-to-e bound.}
Since
\[
E_t=E_{t-1}f_t(p_t),
\]
and \(E_{t-1}\) is \(\mathcal H_{t-1}\)-measurable, we have
\[
\mathbb E(E_t\mid\mathcal H_{t-1},F=0)
=
E_{t-1}\,
\mathbb E(f_t(p_t)\mid\mathcal H_{t-1},F=0).
\]
Thus it is enough to show that
\[
\mathbb E(f_t(p_t)\mid\mathcal H_{t-1},F=0)\le1.
\]
Conditional on \(\mathcal H_{t-1}\), the predictable function \(f_t\) is fixed. Set \(g:=f_t\). Then \(g:[0,1]\to[0,\infty)\) is deterministic, non-increasing, and satisfies
\[
\int_0^1 g(u)\,du
=
\int_0^1 f_t(u)\,du
\le1.
\]

\noindent\textbf{Step 2. Convert monotonicity of \(g\) into a lower-tail event for \(p_t\).}
For each \(y\ge0\), define
\[
A_y:=\{u\in[0,1]:g(u)>y\},
\qquad
a_y:=\lambda(A_y),
\]
where \(\lambda\) denotes Lebesgue measure on \([0,1]\). Since \(g\) is non-increasing, \(A_y\) is an initial segment of \([0,1]\) in the following sense: if \(u\in A_y\) and \(0\le v\le u\), then \(v\in A_y\). Indeed, \(v\le u\) implies \(g(v)\ge g(u)>y\).

This implies that every \(u\in A_y\) satisfies \(u\le a_y\). To see this, suppose \(u\in A_y\) and \(u>a_y\). Since \(A_y\) contains the whole interval \([0,u]\), its Lebesgue measure is at least \(u\), contradicting \(a_y=\lambda(A_y)<u\). Therefore,
\[
\{g(p_t)>y\}
=
\{p_t\in A_y\}
\subseteq
\{p_t\le a_y\}.
\]

\noindent\textbf{Step 3. Use conditional super-uniformity.}
By the inclusion from Step 2,
\[
\mathbb P(g(p_t)>y\mid\mathcal H_{t-1},F=0)
\le
\mathbb P(p_t\le a_y\mid\mathcal H_{t-1},F=0).
\]
Since \(a_y\in[0,1]\), the conditional super-uniformity condition \eqref{eq:su} gives
\[
\mathbb P(p_t\le a_y\mid\mathcal H_{t-1},F=0)
\le
a_y.
\]
Combining the last two displays,
\[
\mathbb P(g(p_t)>y\mid\mathcal H_{t-1},F=0)
\le
a_y.
\]

\noindent\textbf{Step 4. Integrate the tail bound.}
By the layer-cake identity for nonnegative random variables,
\[
\mathbb E(g(p_t)\mid\mathcal H_{t-1},F=0)
=
\int_0^\infty
\mathbb P(g(p_t)>y\mid\mathcal H_{t-1},F=0)\,dy.
\]
Using the bound from Step 3,
\[
\mathbb E(g(p_t)\mid\mathcal H_{t-1},F=0)
\le
\int_0^\infty a_y\,dy.
\]
Now substitute the definition \(a_y=\lambda(A_y)\):
\[
\int_0^\infty a_y\,dy
=
\int_0^\infty
\lambda\{u\in[0,1]:g(u)>y\}
\,dy.
\]
Writing the Lebesgue measure as an integral,
\[
\lambda\{u\in[0,1]:g(u)>y\}
=
\int_0^1 \mathbf 1\{g(u)>y\}\,du.
\]
Substituting this into the preceding display gives
\[
\int_0^\infty a_y\,dy
=
\int_0^\infty
\int_0^1
\mathbf 1\{g(u)>y\}\,du\,dy.
\]
Since the integrand is nonnegative, Tonelli's theorem allows us to switch the order of integration:
\[
\int_0^\infty
\int_0^1
\mathbf 1\{g(u)>y\}\,du\,dy
=
\int_0^1
\int_0^\infty
\mathbf 1\{g(u)>y\}\,dy\,du.
\]
For each fixed \(u\), the inner integral is
\[
\int_0^\infty
\mathbf 1\{g(u)>y\}\,dy
=
g(u),
\]
because the indicator equals one exactly for \(0\le y<g(u)\). Therefore,
\[
\int_0^\infty a_y\,dy
=
\int_0^1 g(u)\,du.
\]
By Step 1, \(\int_0^1 g(u)\,du\le1\). Hence
\[
\mathbb E(g(p_t)\mid\mathcal H_{t-1},F=0)\le1.
\]
Since \(g=f_t\), this is
\[
\mathbb E(f_t(p_t)\mid\mathcal H_{t-1},F=0)\le1.
\]

\noindent\textbf{Step 5. Complete the e-process verification.}
Returning to \(E_t=E_{t-1}f_t(p_t)\), Step 1 and Step 4 give
\[
\mathbb E(E_t\mid\mathcal H_{t-1},F=0)
=
E_{t-1}\,
\mathbb E(f_t(p_t)\mid\mathcal H_{t-1},F=0)
\le
E_{t-1}.
\]
Thus \((E_t)_{t\ge0}\) is a nonnegative supermartingale under the infeasible regime \(F=0\), with \(E_0=1\). Equivalently, it is an e-process for the infeasible regime. This proves Proposition~\ref{prop:betting_martingale}. \qed
}

{\color{black}
\subsection{Proof of Proposition~\ref{prop:always_validity}}
\label{proof:prop_always_validity}

Write \(\mathbb P_0(\cdot):=\mathbb P(\cdot\mid F=0)\). By Proposition~\ref{prop:betting_martingale}, \((E_t)_{t\ge0}\) is a nonnegative supermartingale under \(\mathbb P_0\), with \(E_0=1\).

\noindent\textbf{Step 1. Relate terminal release to threshold crossing.}
Recall that \(\tau_\alpha:=\inf\{t\ge1:E_t\ge1/\alpha\}\) and \(D:=\mathbf 1\{\tau_\alpha\le T_{\max}\}\). Therefore,
\[
\{D=1\}
=
\{\tau_\alpha\le T_{\max}\}.
\]
Since \(\{\tau_\alpha\le T_{\max}\}\subseteq\{\tau_\alpha<\infty\}\), we have
\[
\mathbb P_0(D=1)
\le
\mathbb P_0(\tau_\alpha<\infty).
\]

\noindent\textbf{Step 2. Express threshold crossing as a running supremum event.}
By definition of \(\tau_\alpha\), the event \(\{\tau_\alpha<\infty\}\) occurs exactly when the evidence process crosses \(1/\alpha\) at some finite time. Hence
\[
\{\tau_\alpha<\infty\}
=
\left\{
\sup_{t\ge1}E_t\ge\frac{1}{\alpha}
\right\}.
\]
Thus
\[
\mathbb P_0(\tau_\alpha<\infty)
=
\mathbb P_0\!\left(
\sup_{t\ge1}E_t\ge\frac{1}{\alpha}
\right).
\]

\noindent\textbf{Step 3. Apply Ville's inequality.}
Since \((E_t)_{t\ge0}\) is a nonnegative supermartingale under \(\mathbb P_0\) with \(E_0=1\), Ville's inequality gives
\[
\mathbb P_0\!\left(
\sup_{t\ge1}E_t\ge\frac{1}{\alpha}
\right)
\le
\alpha E_0
=
\alpha.
\]
Combining Steps 1--3,
\[
\mathbb P(D=1\mid F=0)
=
\mathbb P_0(D=1)
\le
\mathbb P_0(\tau_\alpha<\infty)
\le
\alpha.
\]
This proves Proposition~\ref{prop:always_validity}. \qed
}

{\color{black}
\subsection{Proof of Proposition~\ref{prop:observable_separation}}

Let \(O_T=(\mathcal R,H_T)\), and let \((\mathsf O_T,\mathcal A_T)\) denote the measurable space in which \(O_T\) takes values. Since \(D_T\) is measurable with respect to \(\sigma(\mathcal R,H_T)=\sigma(O_T)\), there exists a measurable map \(d_T:\mathsf O_T\to\{0,1\}\) such that \(D_T=d_T(O_T)\) almost surely. Define
\[
A:=\{o\in\mathsf O_T:d_T(o)=1\}.
\]
Then \(A\in\mathcal A_T\).

\noindent\textbf{Step 1. Express release probabilities as probabilities of the same event.}
For \(f\in\{0,1\}\), the conditional law of \(O_T\) given \(F=f\) is \(P_f^T\). Therefore,
\[
\mathbb P(D_T=1\mid F=f)
=
\mathbb P(d_T(O_T)=1\mid F=f)
=
\mathbb P(O_T\in A\mid F=f)
=
P_f^T(A).
\]
In particular,
\[
\left\{
\begin{aligned}
\mathbb P(D_T=1\mid F=1)&=P_1^T(A),\\
\mathbb P(D_T=1\mid F=0)&=P_0^T(A).
\end{aligned}
\right.
\]

\noindent\textbf{Step 2. Apply total variation to the release set.}
By the definition of total variation distance used here,
\[
\TV(P_1^T,P_0^T)
=
\sup_{B\in\mathcal A_T}
\left|P_1^T(B)-P_0^T(B)\right|.
\]
Since \(A\in\mathcal A_T\), taking \(B=A\) in the supremum gives
\[
\left|P_1^T(A)-P_0^T(A)\right|
\le
\TV(P_1^T,P_0^T).
\]
Also,
\[
P_1^T(A)-P_0^T(A)
\le
\left|P_1^T(A)-P_0^T(A)\right|.
\]
Combining the last two displays,
\[
P_1^T(A)-P_0^T(A)
\le
\TV(P_1^T,P_0^T).
\]
Adding \(P_0^T(A)\) to both sides yields
\[
P_1^T(A)
\le
P_0^T(A)+\TV(P_1^T,P_0^T).
\]

\noindent\textbf{Step 3. Translate back to release probabilities.}
Using the identities from Step 1,
\[
\left\{
\begin{aligned}
P_1^T(A)&=\mathbb P(D_T=1\mid F=1),\\
P_0^T(A)&=\mathbb P(D_T=1\mid F=0).
\end{aligned}
\right.
\]
Substituting these identities into the bound from Step 2 gives
\[
\mathbb P(D_T=1\mid F=1)
\le
\mathbb P(D_T=1\mid F=0)
+
\TV(P_1^T,P_0^T).
\]
This proves Proposition~\ref{prop:observable_separation}. \qed

\subsection{Proof of Theorem~\ref{thm:feasible_power_cert_gain}}

Write \(\mathbb P_1(\cdot):=\mathbb P(\cdot\mid F=1)\), \(\mathbb E_1(\cdot):=\mathbb E(\cdot\mid F=1)\), and \(\mathcal H_t:=\widetilde{\mathcal G}_t\). All probabilities and expectations below are taken under \(\mathbb P_1\).

\noindent\textbf{Step 1. Relating terminal release to cumulative calibrated gain.}
Since \(p_t\in[1/(n+1),1]\) and \(f\) is non-increasing, we have
\[
f(1/(n+1))\ge f(p_t)\ge f(1)>0.
\]
Taking logarithms and subtracting \(\log f(1)\) gives
\[
0
\le
\log f(p_t)-\log f(1)
\le
\log f(1/(n+1))-\log f(1).
\]
By the definitions of \(Z_t\) and \(Z_{\max}\), this is \(0\le Z_t\le Z_{\max}\).

Under the fixed calibrator \(f\), the evidence at time \(T\) is \(E_T=\prod_{t=1}^T f(p_t)\). Because every factor is positive,
\[
\log E_T
=
\log\left\{\prod_{t=1}^T f(p_t)\right\}
=
\sum_{t=1}^T \log f(p_t).
\]
Since \(Z_t=\log f(p_t)-\log f(1)\), we have \(\log f(p_t)=\log f(1)+Z_t\). Substituting this identity into the preceding display gives
\[
\log E_T
=
\sum_{t=1}^T\{\log f(1)+Z_t\}
=
T\log f(1)+\sum_{t=1}^T Z_t.
\]
Thus \(E_T\ge1/\alpha\) is equivalent to
\[
T\log f(1)+\sum_{t=1}^T Z_t
\ge
\log(1/\alpha),
\]
which is equivalent, after subtracting \(T\log f(1)\), to
\[
\sum_{t=1}^T Z_t
\ge
\log(1/\alpha)-T\log f(1)
=
A_{\alpha,T}.
\]
Therefore
\[
\{E_T\ge1/\alpha\}
=
\left\{\sum_{t=1}^T Z_t\ge A_{\alpha,T}\right\}.
\]
If \(E_T\ge1/\alpha\), then the evidence process has crossed the release threshold by time \(T\), possibly first crossing exactly at \(T\). Hence
\[
\left\{\sum_{t=1}^T Z_t\ge A_{\alpha,T}\right\}
\subseteq
\{\tau_\alpha\le T\}.
\]
Taking complements gives
\[
\{\tau_\alpha>T\}
\subseteq
\left\{\sum_{t=1}^T Z_t<A_{\alpha,T}\right\},
\]
and therefore
\[
\mathbb P_1(\tau_\alpha>T)
\le
\mathbb P_1\!\left(\sum_{t=1}^T Z_t<A_{\alpha,T}\right).
\]

\noindent\textbf{Step 2. From correct-candidate gain to a lower bound on expected total gain.}
Let \(m_t:=\mathbb E_1(Z_t\mid\mathcal H_{t-1})\). Since \(0\le Z_t\le Z_{\max}\), conditional expectations preserve the bounds, so \(0\le m_t\le Z_{\max}\). Also, since \(Y_t\in\{0,1\}\) and \(Z_t\ge0\), we have \(0\le Y_tZ_t\le Z_t\). Taking conditional expectations given \(\mathcal H_{t-1}\) gives
\[
\mathbb E_1(Y_tZ_t\mid\mathcal H_{t-1})
\le
\mathbb E_1(Z_t\mid\mathcal H_{t-1})
=
m_t.
\]
Summing this inequality over \(t=1,\ldots,T\) yields
\[
\sum_{t=1}^T
\mathbb E_1(Y_tZ_t\mid\mathcal H_{t-1})
\le
\sum_{t=1}^T m_t.
\]
By Assumption~\ref{ass:certifiable_gain},
\[
\sum_{t=1}^T
\mathbb E_1(Y_tZ_t\mid\mathcal H_{t-1})
\ge
B_T
\qquad\text{almost surely}.
\]
Combining the last two displays gives
\[
\sum_{t=1}^T m_t\ge B_T
\qquad\text{almost surely}.
\]

\noindent\textbf{Step 3. Martingale-difference decomposition.}
Define \(X_t:=Z_t-m_t\). Since \(Z_t\) is \(\mathcal H_t\)-measurable and \(m_t\) is \(\mathcal H_{t-1}\)-measurable, \(X_t\) is \(\mathcal H_t\)-measurable. Moreover,
\[
\mathbb E_1(X_t\mid\mathcal H_{t-1})
=
\mathbb E_1(Z_t-m_t\mid\mathcal H_{t-1}).
\]
By linearity of conditional expectation,
\[
\mathbb E_1(Z_t-m_t\mid\mathcal H_{t-1})
=
\mathbb E_1(Z_t\mid\mathcal H_{t-1})
-
\mathbb E_1(m_t\mid\mathcal H_{t-1}).
\]
Since \(m_t\) is \(\mathcal H_{t-1}\)-measurable, \(\mathbb E_1(m_t\mid\mathcal H_{t-1})=m_t\). Since \(m_t=\mathbb E_1(Z_t\mid\mathcal H_{t-1})\), we get
\[
\mathbb E_1(X_t\mid\mathcal H_{t-1})=m_t-m_t=0.
\]
Thus \(M_s:=\sum_{t=1}^s X_t\), \(s=0,\ldots,T\), is a martingale.

The bounds \(0\le Z_t\le Z_{\max}\) imply
\[
-m_t\le Z_t-m_t\le Z_{\max}-m_t,
\]
or equivalently,
\[
-m_t\le X_t\le Z_{\max}-m_t.
\]
The conditional range length of \(X_t\) is
\[
(Z_{\max}-m_t)-(-m_t)=Z_{\max}.
\]

\noindent\textbf{Step 4. Conditional exponential bound for the martingale differences.}
Fix \(\lambda>0\) and \(t\). Conditional on \(\mathcal H_{t-1}\), let \(a_t:=-m_t\), \(b_t:=Z_{\max}-m_t\), and \(c:=b_t-a_t=Z_{\max}\). Then \(X_t\in[a_t,b_t]\) and \(\mathbb E_1(X_t\mid\mathcal H_{t-1})=0\).

For any \(x\in[a_t,b_t]\), convexity of \(x\mapsto e^{-\lambda x}\) gives the linear interpolation bound
\[
e^{-\lambda x}
\le
\frac{b_t-x}{b_t-a_t}e^{-\lambda a_t}
+
\frac{x-a_t}{b_t-a_t}e^{-\lambda b_t}.
\]
Substituting \(x=X_t\) and taking conditional expectations,
\[
\mathbb E_1(e^{-\lambda X_t}\mid\mathcal H_{t-1})
\le
\frac{e^{-\lambda a_t}}{b_t-a_t}
\mathbb E_1(b_t-X_t\mid\mathcal H_{t-1})
+
\frac{e^{-\lambda b_t}}{b_t-a_t}
\mathbb E_1(X_t-a_t\mid\mathcal H_{t-1}).
\]
Because \(a_t\) and \(b_t\) are \(\mathcal H_{t-1}\)-measurable and \(\mathbb E_1(X_t\mid\mathcal H_{t-1})=0\), we have
\[
\mathbb E_1(b_t-X_t\mid\mathcal H_{t-1})=b_t
\quad\text{and}\quad
\mathbb E_1(X_t-a_t\mid\mathcal H_{t-1})=-a_t.
\]
Using \(b_t-a_t=c\), the previous display becomes
\[
\mathbb E_1(e^{-\lambda X_t}\mid\mathcal H_{t-1})
\le
\frac{b_t}{c}e^{-\lambda a_t}
+
\frac{-a_t}{c}e^{-\lambda b_t}.
\]
Let \(q:=-a_t/c=m_t/Z_{\max}\). Since \(0\le m_t\le Z_{\max}\), we have \(q\in[0,1]\). Also \(a_t=-qc\), \(b_t=(1-q)c\), \(b_t/c=1-q\), and \((-a_t)/c=q\). With \(\theta:=\lambda c=\lambda Z_{\max}\), this gives
\[
\mathbb E_1(e^{-\lambda X_t}\mid\mathcal H_{t-1})
\le
(1-q)e^{q\theta}+q e^{-(1-q)\theta}.
\]

It remains to bound the last expression. Define \(A(\theta):=(1-q)e^{q\theta}+q e^{-(1-q)\theta}\) and \(\psi(\theta):=\log A(\theta)\). Then \(A(0)=1\), so \(\psi(0)=0\). Differentiating \(A\),
\[
A'(\theta)
=
q(1-q)e^{q\theta}
-
q(1-q)e^{-(1-q)\theta}
=
q(1-q)\{e^{q\theta}-e^{-(1-q)\theta}\}.
\]
Thus \(A'(0)=q(1-q)(1-1)=0\), and hence \(\psi'(0)=A'(0)/A(0)=0\).

Now set
\[
w(\theta):=\frac{(1-q)e^{q\theta}}{A(\theta)}.
\]
Then \(1-w(\theta)=q e^{-(1-q)\theta}/A(\theta)\). Since \(\psi'(\theta)=A'(\theta)/A(\theta)\), we have
\[
\psi'(\theta)
=
q\,w(\theta)
-
(1-q)\{1-w(\theta)\}.
\]
Expanding the right-hand side,
\[
q\,w(\theta)
-
(1-q)\{1-w(\theta)\}
=
q\,w(\theta)-(1-q)+(1-q)w(\theta)
=
w(\theta)-(1-q).
\]
Therefore \(\psi'(\theta)=w(\theta)-(1-q)\), and so \(\psi''(\theta)=w'(\theta)\).

To compute \(w'(\theta)\), write \(N(\theta):=(1-q)e^{q\theta}\), so that \(w(\theta)=N(\theta)/A(\theta)\). Since \(N'(\theta)=qN(\theta)\),
\[
w'(\theta)
=
\frac{N'(\theta)A(\theta)-N(\theta)A'(\theta)}{A(\theta)^2}
=
\frac{N(\theta)}{A(\theta)}
\left\{
q-\frac{A'(\theta)}{A(\theta)}
\right\}.
\]
Because \(N(\theta)/A(\theta)=w(\theta)\) and \(A'(\theta)/A(\theta)=\psi'(\theta)=w(\theta)-(1-q)\), this becomes
\[
w'(\theta)
=
w(\theta)\{q-w(\theta)+(1-q)\}
=
w(\theta)\{1-w(\theta)\}.
\]
Thus \(\psi''(\theta)=w(\theta)\{1-w(\theta)\}\). Since \(w(\theta)\in[0,1]\), we have \(w(\theta)\{1-w(\theta)\}\le1/4\), and hence
\[
\psi''(\theta)\le\frac14
\qquad\text{for all }\theta\ge0.
\]
Taylor's formula with integral remainder around \(0\) gives
\[
\psi(\theta)
=
\psi(0)+\theta\psi'(0)+\int_0^\theta(\theta-r)\psi''(r)\,dr.
\]
Using \(\psi(0)=0\), \(\psi'(0)=0\), and \(\psi''(r)\le1/4\),
\[
\psi(\theta)
\le
\int_0^\theta \frac{\theta-r}{4}\,dr
=
\frac14\left[\theta r-\frac{r^2}{2}\right]_{r=0}^{r=\theta}
=
\frac14\left(\theta^2-\frac{\theta^2}{2}\right)
=
\frac{\theta^2}{8}.
\]
Exponentiating, \(A(\theta)=e^{\psi(\theta)}\le e^{\theta^2/8}\). Since \(\theta=\lambda Z_{\max}\), we conclude that
\[
\mathbb E_1(e^{-\lambda X_t}\mid\mathcal H_{t-1})
\le
\exp\!\left\{\frac{\lambda^2 Z_{\max}^2}{8}\right\}.
\]

\noindent\textbf{Step 5. Martingale tail bound and optimization.}
Let \(S_T^X:=\sum_{t=1}^T X_t\). For any \(\lambda>0\),
\[
\mathbb E_1(e^{-\lambda S_T^X})
=
\mathbb E_1\!\left[
\exp\!\left\{-\lambda\sum_{t=1}^{T-1}X_t\right\}
e^{-\lambda X_T}
\right].
\]
By the tower property,
\[
\mathbb E_1(e^{-\lambda S_T^X})
=
\mathbb E_1\!\left[
\exp\!\left\{-\lambda\sum_{t=1}^{T-1}X_t\right\}
\mathbb E_1(e^{-\lambda X_T}\mid\mathcal H_{T-1})
\right].
\]
Using the conditional exponential bound from Step 4,
\[
\mathbb E_1(e^{-\lambda S_T^X})
\le
\exp\!\left\{\frac{\lambda^2 Z_{\max}^2}{8}\right\}
\mathbb E_1\!\left[
\exp\!\left\{-\lambda\sum_{t=1}^{T-1}X_t\right\}
\right].
\]
Repeating this argument for \(T-1,T-2,\ldots,1\) yields
\[
\mathbb E_1(e^{-\lambda S_T^X})
\le
\exp\!\left\{\frac{T\lambda^2 Z_{\max}^2}{8}\right\}.
\]

Let \(\Delta_{\alpha,T}=B_T-A_{\alpha,T}>0\). By Markov's inequality,
\[
\mathbb P_1(S_T^X\le-\Delta_{\alpha,T})
=
\mathbb P_1(e^{-\lambda S_T^X}\ge e^{\lambda\Delta_{\alpha,T}})
\le
e^{-\lambda\Delta_{\alpha,T}}
\mathbb E_1(e^{-\lambda S_T^X}).
\]
Using the moment bound above,
\[
\mathbb P_1(S_T^X\le-\Delta_{\alpha,T})
\le
\exp\!\left\{
-\lambda\Delta_{\alpha,T}
+
\frac{T\lambda^2 Z_{\max}^2}{8}
\right\}.
\]
To minimize the exponent, define \(\phi(\lambda):=-\lambda\Delta_{\alpha,T}+T\lambda^2Z_{\max}^2/8\). Then
\[
\phi'(\lambda)
=
-\Delta_{\alpha,T}
+
\frac{T Z_{\max}^2}{4}\lambda.
\]
Solving \(\phi'(\lambda)=0\) gives
\[
\lambda_*
=
\frac{4\Delta_{\alpha,T}}{T Z_{\max}^2}.
\]
Because \(\Delta_{\alpha,T}>0\) and \(Z_{\max}>0\), we have \(\lambda_*>0\). Also \(\phi''(\lambda)=T Z_{\max}^2/4>0\), so \(\lambda_*\) is the minimizer.

Substituting \(\lambda_*\) into the exponent,
\[
-\lambda_*\Delta_{\alpha,T}
=
-
\frac{4\Delta_{\alpha,T}}{T Z_{\max}^2}\Delta_{\alpha,T}
=
-
\frac{4\Delta_{\alpha,T}^2}{T Z_{\max}^2},
\]
and
\[
\frac{T\lambda_*^2 Z_{\max}^2}{8}
=
\frac{T Z_{\max}^2}{8}
\left(
\frac{4\Delta_{\alpha,T}}{T Z_{\max}^2}
\right)^2
=
\frac{T Z_{\max}^2}{8}
\cdot
\frac{16\Delta_{\alpha,T}^2}{T^2 Z_{\max}^4}
=
\frac{2\Delta_{\alpha,T}^2}{T Z_{\max}^2}.
\]
Therefore
\[
-\lambda_*\Delta_{\alpha,T}
+
\frac{T\lambda_*^2 Z_{\max}^2}{8}
=
-
\frac{4\Delta_{\alpha,T}^2}{T Z_{\max}^2}
+
\frac{2\Delta_{\alpha,T}^2}{T Z_{\max}^2}
=
-
\frac{2\Delta_{\alpha,T}^2}{T Z_{\max}^2}.
\]
Hence
\[
\mathbb P_1(S_T^X\le-\Delta_{\alpha,T})
\le
\exp\!\left\{
-\frac{2\Delta_{\alpha,T}^2}{T Z_{\max}^2}
\right\}.
\]

\noindent\textbf{Step 6. Completing the release-probability bound.}
By definition of \(X_t\),
\[
S_T^X
=
\sum_{t=1}^T X_t
=
\sum_{t=1}^T(Z_t-m_t)
=
\sum_{t=1}^T Z_t-\sum_{t=1}^T m_t.
\]
On the event \(\{\sum_{t=1}^T Z_t<A_{\alpha,T}\}\), and using \(\sum_{t=1}^T m_t\ge B_T\) from Step 2, we have
\[
S_T^X
=
\sum_{t=1}^T Z_t-\sum_{t=1}^T m_t
<
A_{\alpha,T}-B_T.
\]
Since \(\Delta_{\alpha,T}=B_T-A_{\alpha,T}\), we have \(A_{\alpha,T}-B_T=-\Delta_{\alpha,T}\). Therefore
\[
\left\{\sum_{t=1}^T Z_t<A_{\alpha,T}\right\}
\subseteq
\{S_T^X<-\Delta_{\alpha,T}\}
\subseteq
\{S_T^X\le-\Delta_{\alpha,T}\}.
\]
Taking probabilities and applying Step 5 gives
\[
\mathbb P_1\!\left(\sum_{t=1}^T Z_t<A_{\alpha,T}\right)
\le
\mathbb P_1(S_T^X\le-\Delta_{\alpha,T})
\le
\exp\!\left\{
-\frac{2\Delta_{\alpha,T}^2}{T Z_{\max}^2}
\right\}.
\]
Combining this with the bound from Step 1,
\[
\mathbb P_1(\tau_\alpha>T)
\le
\mathbb P_1\!\left(\sum_{t=1}^T Z_t<A_{\alpha,T}\right),
\]
we obtain
\[
\mathbb P_1(\tau_\alpha>T)
\le
\exp\!\left\{
-\frac{2\Delta_{\alpha,T}^2}{T Z_{\max}^2}
\right\}.
\]
Taking complements,
\[
\mathbb P_1(\tau_\alpha\le T)
=
1-\mathbb P_1(\tau_\alpha>T)
\ge
1-
\exp\!\left\{
-\frac{2\Delta_{\alpha,T}^2}{T Z_{\max}^2}
\right\}.
\]
Finally, substituting back \(\mathbb P_1(\cdot)=\mathbb P(\cdot\mid F=1)\),
\[
\mathbb P(\tau_\alpha\le T\mid F=1)
\ge
1-
\exp\!\left\{
-\frac{2\Delta_{\alpha,T}^2}{T Z_{\max}^2}
\right\}.
\]
This proves Theorem~\ref{thm:feasible_power_cert_gain}. \qed

\subsection{Proof of Theorem~\ref{thm:drift_robustness}}

Fix \(T\in\mathbb N\). Throughout the proof, we work under the conditional law given \(\sigma(\mathcal R)\) and on the event \(\{F=0\}\). Write \(\mathbb P_{\mathcal R,0}\) and \(\mathbb E_{\mathcal R,0}\) for probability and expectation under this conditional law, let \(\mathcal H_t:=\widetilde{\mathcal G}_t\), and set \(\beta:=1+M\epsilon\). We work on the probability-one event on which \(\TV(Q_{t,0},P_0)\le\epsilon\) for \(t=1,\ldots,T\).

\noindent\textbf{Step 1. One-step inflation bound.}
Since \(e_t=f(p_t)\) and \(Q_{t,0}=\mathcal L(p_t\mid\mathcal H_{t-1},F=0)\),
\[
\mathbb E_{\mathcal R,0}(e_t\mid\mathcal H_{t-1})
=
\mathbb E_{\mathcal R,0}(f(p_t)\mid\mathcal H_{t-1})
=
\int_0^1 f(u)\,Q_{t,0}(du).
\]
We next bound this integral by the nominal law \(P_0\). For any probability measure \(Q\) on \([0,1]\) and any measurable \(g:[0,1]\to[0,M]\), the layer-cake representation gives
\[
\int_0^1 g(u)\,Q(du)
=
\int_0^M Q\{g>r\}\,dr
\]
and
\[
\int_0^1 g(u)\,P_0(du)
=
\int_0^M P_0\{g>r\}\,dr.
\]
Subtracting the second display from the first gives
\[
\int_0^1 g(u)\,Q(du)
-
\int_0^1 g(u)\,P_0(du)
=
\int_0^M
\bigl[
Q\{g>r\}-P_0\{g>r\}
\bigr]\,dr.
\]
For each \(r\), the set \(\{u:g(u)>r\}\) is measurable, so
\[
Q\{g>r\}-P_0\{g>r\}
\le
\TV(Q,P_0).
\]
Therefore,
\[
\int_0^1 g(u)\,Q(du)
-
\int_0^1 g(u)\,P_0(du)
\le
\int_0^M \TV(Q,P_0)\,dr.
\]
Since \(\TV(Q,P_0)\) does not depend on \(r\),
\[
\int_0^M \TV(Q,P_0)\,dr
=
M\TV(Q,P_0).
\]
Hence
\[
\int_0^1 g(u)\,Q(du)
\le
\int_0^1 g(u)\,P_0(du)
+
M\TV(Q,P_0).
\]
Applying this inequality with \(g=f\) and \(Q=Q_{t,0}\), we obtain
\[
\int_0^1 f(u)\,Q_{t,0}(du)
\le
\int_0^1 f(u)\,P_0(du)
+
M\TV(Q_{t,0},P_0).
\]
By the theorem assumptions, \(\int_0^1 f(u)\,P_0(du)\le1\) and \(\TV(Q_{t,0},P_0)\le\epsilon\). Substituting these two bounds gives
\[
\int_0^1 f(u)\,Q_{t,0}(du)
\le
1+M\epsilon.
\]
Since \(\beta=1+M\epsilon\), we conclude that
\[
\mathbb E_{\mathcal R,0}(e_t\mid\mathcal H_{t-1})
\le
\beta.
\]

\noindent\textbf{Step 2. Deflating the evidence process.}
Define the deflated process \(\widetilde E_t:=\beta^{-t}E_t\) for \(t\ge0\). Since \(E_t=E_{t-1}e_t\),
\[
\widetilde E_t
=
\beta^{-t}E_t
=
\beta^{-t}E_{t-1}e_t.
\]
Taking conditional expectations given \(\mathcal H_{t-1}\),
\[
\mathbb E_{\mathcal R,0}(\widetilde E_t\mid\mathcal H_{t-1})
=
\mathbb E_{\mathcal R,0}(\beta^{-t}E_{t-1}e_t\mid\mathcal H_{t-1}).
\]
Because \(\beta^{-t}\) is deterministic and \(E_{t-1}\) is \(\mathcal H_{t-1}\)-measurable,
\[
\mathbb E_{\mathcal R,0}(\widetilde E_t\mid\mathcal H_{t-1})
=
\beta^{-t}E_{t-1}
\mathbb E_{\mathcal R,0}(e_t\mid\mathcal H_{t-1}).
\]
Using the one-step bound from Step 1,
\[
\mathbb E_{\mathcal R,0}(\widetilde E_t\mid\mathcal H_{t-1})
\le
\beta^{-t}E_{t-1}\beta.
\]
Since \(\beta^{-t}\beta=\beta^{-(t-1)}\),
\[
\beta^{-t}E_{t-1}\beta
=
\beta^{-(t-1)}E_{t-1}.
\]
By definition, \(\widetilde E_{t-1}=\beta^{-(t-1)}E_{t-1}\). Therefore,
\[
\mathbb E_{\mathcal R,0}(\widetilde E_t\mid\mathcal H_{t-1})
\le
\widetilde E_{t-1}.
\]
Thus \((\widetilde E_t)_{t\ge0}\) is a nonnegative supermartingale under \(\mathbb P_{\mathcal R,0}\). Also, since \(E_0=1\) and \(\beta^0=1\), we have \(\widetilde E_0=1\).

\noindent\textbf{Step 3. Relating release to the deflated process.}
If \(\tau_\alpha\le T\), then by definition of \(\tau_\alpha\), there exists a time \(t\in\{1,\ldots,T\}\) such that \(E_t\ge1/\alpha\). For this same \(t\),
\[
\widetilde E_t
=
\beta^{-t}E_t
\ge
\beta^{-t}\alpha^{-1}.
\]
Since \(\epsilon\ge0\) and \(M\ge0\), we have \(\beta=1+M\epsilon\ge1\). Because \(t\le T\) and \(\beta\ge1\),
\[
\beta^{-t}\ge\beta^{-T}.
\]
Therefore,
\[
\widetilde E_t
\ge
\beta^{-T}\alpha^{-1}.
\]
This proves the event inclusion
\[
\{\tau_\alpha\le T\}
\subseteq
\left\{
\sup_{1\le t\le T}\widetilde E_t
\ge
\beta^{-T}\alpha^{-1}
\right\}.
\]

\noindent\textbf{Step 4. Applying Ville's inequality.}
By Ville's inequality for the nonnegative supermartingale \((\widetilde E_t)_{t\ge0}\),
\[
\mathbb P_{\mathcal R,0}\!\left(
\sup_{1\le t\le T}\widetilde E_t
\ge
\beta^{-T}\alpha^{-1}
\right)
\le
\frac{\mathbb E_{\mathcal R,0}(\widetilde E_0)}
{\beta^{-T}\alpha^{-1}}.
\]
Since \(\widetilde E_0=1\),
\[
\frac{\mathbb E_{\mathcal R,0}(\widetilde E_0)}
{\beta^{-T}\alpha^{-1}}
=
\frac{1}{\beta^{-T}\alpha^{-1}}.
\]
Simplifying the denominator,
\[
\frac{1}{\beta^{-T}\alpha^{-1}}
=
\beta^T\alpha.
\]
Combining this with the event inclusion from Step 3 gives
\[
\mathbb P_{\mathcal R,0}(\tau_\alpha\le T)
\le
\alpha\beta^T.
\]
Finally, substituting \(\beta=1+M\epsilon\),
\[
\alpha\beta^T
=
\alpha(1+M\epsilon)^T.
\]
Hence
\[
\mathbb P_{\mathcal R,0}(\tau_\alpha\le T)
\le
\alpha(1+M\epsilon)^T.
\]
Restoring the conditioning notation, this is
\[
\mathbb P\!\left(\tau_\alpha\le T\,\middle|\,\sigma(\mathcal R),F=0\right)
\le
\alpha(1+M\epsilon)^T.
\]
This proves Theorem~\ref{thm:drift_robustness}. \qed
}

\end{document}